\documentclass{article}
\usepackage[letterpaper,margin=1in]{geometry}
\usepackage[OT1]{fontenc}
\usepackage{amsmath,amsfonts}
\usepackage{algorithmic}
\usepackage{algorithm}
\usepackage{array}
\usepackage[caption=false,font=normalsize,labelfont=sf,textfont=sf]{subfig}
\usepackage{float}
\usepackage{url}
\usepackage{verbatim}
\usepackage{graphicx}
\usepackage{makecell}

\usepackage{enumitem}
\usepackage{xcolor}    
\usepackage{array}    
\usepackage{colortbl}   
\usepackage{multirow}   
\usepackage{booktabs}
\usepackage{lineno}
\usepackage{amssymb}
\usepackage{pdflscape} 
\usepackage{afterpage}

\usepackage{tabularx}
\usepackage{adjustbox}
\newcolumntype{C}{>{\centering\arraybackslash}X}
\usepackage{hyperref}
\hypersetup{
    citecolor=blue,     
    linkcolor=red,      
    urlcolor=blue,     
    colorlinks=true     
}
\usepackage{cleveref}
\usepackage{titling}
\thanksmarkseries{arabic}
\usepackage[symbol]{footmisc}
\begin{document}

\title{\LARGE LatticeWorld: 
	A Multimodal Large Language Model-Empowered Framework for Interactive Complex World Generation
}

\author{Yinglin Duan{\color{blue}${}^*$}\thanks{NetEase, Inc., China. 
		\texttt{duanyinglin@corp.netease.com}} 
	\quad 
	Zhengxia Zou{\color{blue}${}^*$}\thanks{Beihang University, China. 
		\texttt{zhengxiazou@buaa.edu.cn}}
	\quad
	Tongwei Gu{\color{blue}${}^*$}\thanks{NetEase, Inc., China. 
		\texttt{gutongwei@corp.netease.com}}
	\quad
	Wei Jia\thanks{NetEase, Inc., China. 
		\texttt{jiawei@corp.netease.com}}
	\quad  
	Zhan Zhao\thanks{NetEase, Inc., China. 
		\texttt{zhaozhan03@corp.netease.com}}
	\quad \\
	Luyi Xu\thanks{Work done while at NetEase, Inc., China. 
		\texttt{lucxu@vip.163.com}}
	\quad
	Xinzhu Liu\thanks{Tsinghua University, China. 
		\texttt{liuxz\_cs@163.com}}
	\quad
	Yenan Lin\thanks{Independent Researcher \& Technical Artists, China. 
		\texttt{yenan\_lin@foxmail.com}}
	\quad
	Hao Jiang\thanks{NetEase, Inc., China. 
		\texttt{jianghao06@corp.netease.com}}
	\quad
	Kang Chen{\color{blue}${}^\dag$}\thanks{NetEase, Inc., China. 
		\texttt{ckn6763@corp.netease.com}}
	\quad
	Shuang Qiu{\color{blue}${}^\dag$}\thanks{City University of Hong Kong, China. 
		\texttt{shuanqiu@cityu.edu.hk}} 
}

\date{\today}
\footnotetext[1]{Equal Contribution.}
\footnotetext[2]{Corresponding Authors.}
\maketitle

\begin{figure*}[t]
\centering
\includegraphics[width=\linewidth]{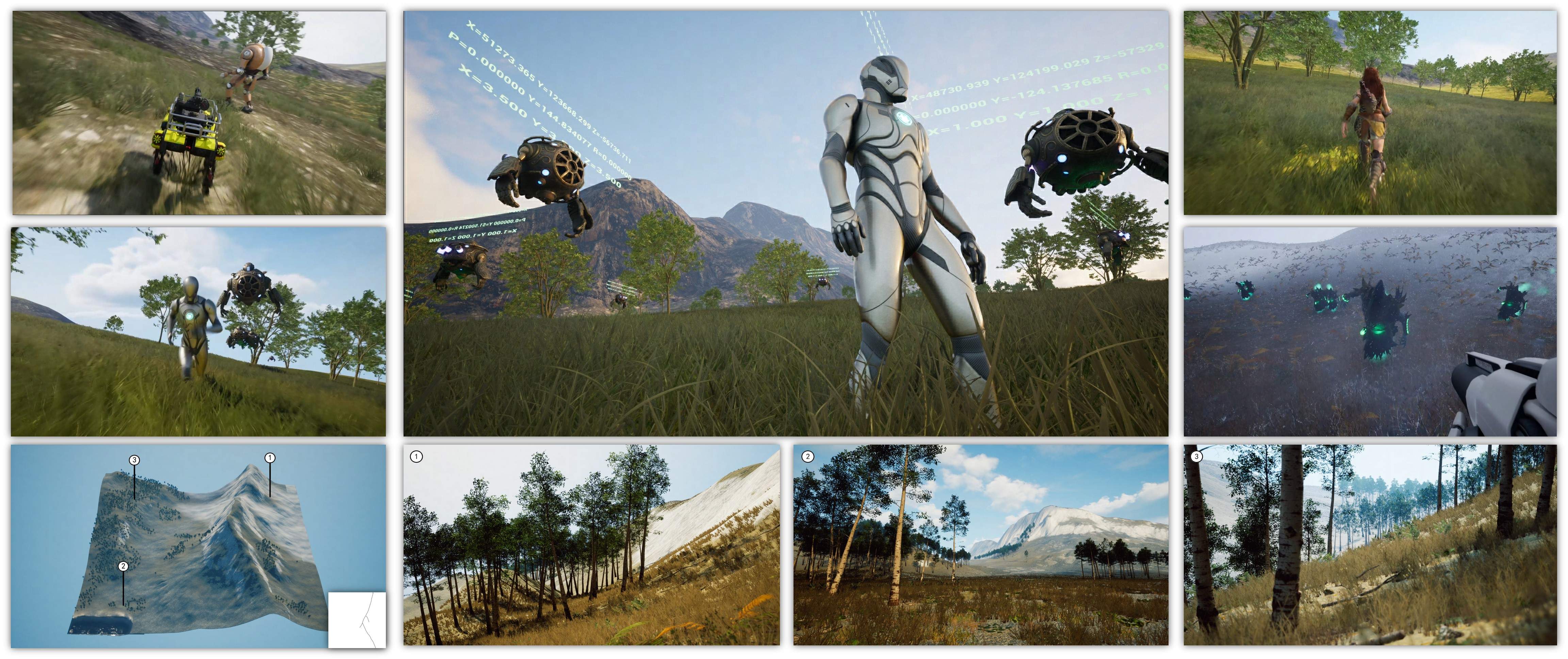}
\caption{Demonstration of generated results via LatticeWorld
}
\label{fig:teaser}
\end{figure*}


\begin{abstract}
Recent research has been increasingly focusing on developing 3D world models that simulate complex real-world scenarios. World models have found broad applications across various domains, including embodied AI, autonomous driving, entertainment, etc. A more realistic simulation with accurate physics will effectively narrow the sim-to-real gap and allow us to gather rich information about the real world conveniently. While traditional manual modeling has enabled the creation of virtual 3D scenes, modern approaches have leveraged advanced machine learning algorithms for 3D world generation, with most recent advances focusing on generative methods that can create virtual worlds based on user instructions. This work explores such a research direction by proposing LatticeWorld, a simple yet effective 3D world generation framework that streamlines the industrial production pipeline of 3D environments. LatticeWorld leverages lightweight LLMs (LLaMA-2-7B) alongside the industry-grade rendering engine (e.g., Unreal Engine 5) to generate a dynamic environment. Our proposed framework accepts textual descriptions and visual instructions as multimodal inputs and creates large-scale 3D interactive worlds with dynamic agents, featuring competitive multi-agent interaction, high-fidelity physics simulation, and real-time rendering. We conduct comprehensive experiments to evaluate LatticeWorld, showing that it achieves superior accuracy in scene layout generation and visual fidelity. Moreover, LatticeWorld achieves over a $90\times$ increase in industrial production efficiency while maintaining high creative quality compared with traditional manual production methods. Our demo video is available at \url{https://youtu.be/8VWZXpERR18}
\end{abstract}

\section{Introduction}
The development of interactive world models has emerged as a crucial research area in recent studies~\cite{worldlabs, genie2, lu2024genexgeneratingexplorableworld}. These virtual environments aim to simulate real-world complex scenarios, enabling researchers to gather rich observations through extensive and meaningful interactions. Given that obtaining real-world data is typically challenging and expensive, the abundant data generated by world models can be leveraged to train various artificial intelligence (AI) algorithms~\cite{shang2024urbanworld, zhong2024unrealzooenrichingphotorealisticvirtual}, especially those requiring high sample complexity. Furthermore, world models are particularly valuable for learning problems that involve safety and risk considerations. World models have been widely applied across numerous applications, including embodied artificial intelligence, sequential decision-making, autonomous driving, entertainment (e.g., game development, movie production), etc. In essence, world models will enable intelligent systems to acquire spatial intelligence through extensive interactions.

The empirical significance of world models necessitates developing high-fidelity 3D virtual environments, aiming to minimize the sim-to-real gap between simulation and reality. By doing so, we can generate rich data that better aligns with real-world samples, significantly mitigating the learning errors that arise from the sim-to-real gap. 
The development of such high-fidelity virtual environments fundamentally requires the study of 3D scene generation techniques.
Early efforts in this field began with basic manual scene modeling, primarily in applications like game development, which largely relied on manual crafting by artists with high labor costs. To enhance  production efficiency, procedural content generation (PCG) has become  widely used in computer graphics for the automatic creation of 3D virtual environments through algorithms and predefined rules \cite{togelius2011procedural,beukman2022procedural, gambi2019automatically,ayubi2020deterministic,earle2022illuminating,wu2021procedural}. With the advancement of deep learning, there has been extensive research on integrating PCG with deep neural networks \cite{perez2019general, khalifa2020pcgrl,liu2021deep}.
Recently, researchers are particularly focused on using generative models to create content based on user instructions. Neural rendering methods represent an important line of research, such as Score Distillation Sampling with Neural Radiance Fields~\cite{mildenhall2021nerf, muller2022instant,song2023roomdreamer,text2room,lu2024urban,zhang20243d,scenescape,text2nerf} and 3D Gaussian Splatting~\cite{kerbl3Dgaussians,yi2024gaussiandreamer,chen2024text,zhou2024dreamscene360,li2025dreamscene}. However, these approaches lack interactive capabilities, which constrains their practical applications. Another line of research, developed based on diffusion models \cite{fridman2024scenescape,li2024director3d,genie2,lu2024genexgeneratingexplorableworld,deng2023citygeninfinitecontrollable3d}, provides vision-based (e.g., images or videos) solutions for scene generation. Moreover, there have been many works focusing on integrating generative models with 3D modeling platforms\cite{sun20233d,zhou2024scenex,hu2024scenecraft,deng2024citycraft,shang2024urbanworld,wu2024metaurban,xu2023urban}, especially Blender.

Our work advances this field by proposing a novel framework named LatticeWorld,  
a multimodal \textbf{\underline{l}}arge l\textbf{\underline{a}}nguage model-empowered framework for in\textbf{\underline{t}}erac\textbf{\underline{t}}\textbf{\underline{i}}ve \textbf{\underline{c}}ompl\textbf{\underline{e}}x \textbf{\underline{world}} generation. 
Drawing inspiration from standard computer graphics (CG) solutions in industry, our method integrates with industrial PCG production pipelines. LatticeWorld is a simple yet effective framework that seamlessly integrates multimodal LLM with an industry-grade CG rendering engine, the Unreal Engine (UE), distinguishing it from prior works. Compared with Blender, Unreal Engine offers more realistic physics simulation, native multi-agent interaction capabilities, and real-time rendering optimized for interactive experiences. Hence, LatticeWorld inherits UE's distinctive advantages and extended functionality through established plugins. Specifically, LatticeWorld accepts both textual descriptions of the virtual world and visual instructions for terrain elevation (such as height maps or sketches) as inputs. Then, leveraging LLMs' capabilities in symbolic understanding and structured sequence generation, the well-trained multimodal LLMs in our framework generate a carefully defined symbolic representation (matrix) of the scene layout and extract semantically clear configurations of the environment from the input, showcasing excellent interpretability and semantic precision. The rendering engine processes these generated results alongside the visual information to create a large-scale dynamic virtual world populated with multiple interactive agents. The visual condition can guarantee a direct and efficient understanding of elevation in the scene by the LLM, thus leading to accurate generation. 
Our framework eventually generates a playable virtual world, where users can control the main character to interact with other agents exhibiting adversarial behaviors. Thus, LatticeWorld has the capability of creating a competitive environment based on user instructions for AI agent training. Notably, LatticeWorld's multimodal LLMs are built upon the lightweight LLaMA-2-7B model \cite{touvron2023llama2}, demonstrating the potential for achieving sophisticated spatial understanding using smaller-scale LLMs.
We summarize the main contributions of our work as follows:
\begin{itemize}
	\item We propose a simple yet effective 3D complex world generation framework, LatticeWorld, by exploring lightweight LLMs' abilities in spatial understanding and structured sequence generation. 
	LatticeWorld leverages multimodal LLMs alongside the industry-grade rendering engine, UE, to create a dynamic environment, featuring several key advantages: \textbf{(1)} multimodal input, \textbf{(2)} interpretable intermediate representation, \textbf{(3)} realistic physics modeling, \textbf{(4)} dynamic multi-agent interaction, \textbf{(5)} real-time large-scale simulation. Our method is general and can be adapted to other powerful engines such as Unity.

	\item To train the multimodal LLMs in our framework, we propose to construct multimodal datasets that incorporate diverse textual descriptions, height maps, symbolic layout representations, and corresponding environment configurations. During the dataset construction process, we leverage GPT-4o for data annotation along with sophisticated prompt engineering, ensuring both annotation efficiency and accuracy.

	\item We conduct comprehensive experiments to evaluate LatticeWorld's performance, comparing it with existing methods in both layout generation and final environment generation. Our results demonstrate that LatticeWorld achieves superior performance in terms of generation accuracy and visual fidelity across various users' instructions. Additionally, compared with traditional manual production methods in industry, LatticeWorld can achieve over a $90 \times$ increase in industrial production efficiency while maintaining high creative quality.  
\end{itemize}

\section{Related Work}
\noindent \textbf{Procedural Content Generation}. PCG frameworks automate the creation of assets and environments through algorithmic approaches~\cite{togelius2011procedural}, traditionally relying on rule-based systems and parametric models to generate diverse content such as terrains, levels, and gameplay elements~\cite{freiknecht2017survey,  gambi2019automatically,ayubi2020deterministic,wu2021procedural,earle2022illuminating,beukman2022procedural}. These approaches have been widely adopted in industrial pipelines, particularly in game development and virtual environment creation, due to their efficiency in generating large-scale content with controlled variability. Furthermore, many works have explored integrating PCG with AI methods~\cite{liu2021deep, perez2019general, khalifa2020pcgrl} and 3D scene generation ~\cite{sun20233d,raistrick2023infinite,hu2024scenecraft,zhou2024scenex,liu2024controllable}. 

\vspace{5pt}
\noindent\textbf{Neural Rendering for 3D Scene Generation}. 
The field has evolved from Score Distillation Sampling with Neural Radiance Fields (NeRFs)~\cite{mildenhall2021nerf, muller2022instant,song2023roomdreamer,text2room,lu2024urban,zhang20243d, scenescape, text2nerf} to feed-forward architectures~\cite{li2024director3d,fridman2024scenescape}. Recent developments in 3D Gaussian Splatting~\cite{kerbl3Dgaussians,yi2024gaussiandreamer,chen2024text,zhou2024dreamscene360,li2025dreamscene} have enhanced generation efficiency and scene coherence. While these neural rendering approaches excel in visual fidelity, they primarily focus on static content generation and usually lack interactive capabilities, limiting their application in dynamic world model scenarios.

\vspace{5pt}
\noindent\textbf{Vision-Based Interactive World Generation}. 
Diffusion-based approaches have established frameworks for interactive 3D environment creation~\cite{fridman2024scenescape,li2024director3d,deng2023citygeninfinitecontrollable3d}. Recent works like Genie-2~\cite{genie2}, WorldLabs~\cite{worldlabs}, and GenEx~\cite{lu2024genexgeneratingexplorableworld} use diffusion models with image/video inputs to create explorable environments. These methods achieve interactivity through visual prediction but are constrained by the limitations of vision-based simulation.

\vspace{5pt}
\noindent\textbf{Platform-Based Environment Creation}.
In addition to the above methods, recent efforts have focused on integrating generative capabilities with established 3D content creation platforms, each offering distinct computational paradigms and interaction capabilities.
Recent works have demonstrated integration with Blender~\cite{sun20233d,zhou2024scenex,hu2024scenecraft,gd3kr_BlenderGPT_2023,deng2024citycraft,shang2024urbanworld,wu2024metaurban,xu2023urban}, leveraging its comprehensive modeling capabilities and Python scripting ecosystem. These methods primarily emphasize content creation and offline rendering rather than real-time interaction. Although Blender supports real-time rendering through Eevee and physics simulation, these capabilities remain limited for dynamic multi-agent scenarios, and its typical workflow involves content creation followed by export to other platforms, resulting in a separation between generation and interaction phases. Additionally, more specialized simulation platforms like Nvidia Isaac Sim have been utilized for robot interaction tasks~\cite{ren2024infiniteworldunifiedscalablesimulation}, while simulation environments have evolved from symbolic reasoning~\cite{puig2018virtualhome,shridhar2020alfred} to sophisticated physics-based platforms~\cite{szot2022habitat20traininghome,li2021igibson20objectcentricsimulation,lin2021softgymbenchmarkingdeepreinforcement,puig2023habitat30cohabitathumans, IsaacSim}. However, these approaches typically leverage static content rather than enabling dynamic content generation that responds to agent interactions in real-time, such as trees falling due to collision impact.
Our approach leverages Unreal Engine, which offers a fundamentally different paradigm optimized for real-time interactive applications. Unlike Blender, Unreal Engine offers advanced physics simulation, native multi-agent interaction capabilities, and real-time rendering optimized for interactive experiences. This allows us to integrate LLM generative capabilities with responsive virtual environments that support real-time scene modification and agent interaction, making our approach suitable for both creative applications and AI agent training.

\vspace{5pt}
\noindent \textbf{Large Language Models}.
LLMs have demonstrated capabilities in understanding textual instructions, as exemplified by GPT series \cite{brown2020language,openai2024gpt4technicalreport}, LLaMA family \cite{touvron2023llama2,llama3_model}, Mistral \cite{jiang2023mistral}, and task-specific models like DeepSeek-R1 \cite{deepseekai2025deepseekr1incentivizingreasoningcapability}. While these models primarily process unimodal text inputs, recent advances expand their versatility through multimodal integration (e.g., GPT-4 variants \cite{openai2023gpt4v,openai2024gpt4o}, Qwen2-VL\cite{wang2024qwen2}, Claude 3 \cite{anthropic2024claude}, Gemini \cite{geminiteam2024geminifamilyhighlycapable}). The emergence of instruction tuning \cite{raffel2020exploring,wei2022finetuned,ouyang2022training} has empowered LLMs to generalize across unseen tasks through textual instruction comprehension. To extend LLMs' capabilities to visual inputs, frameworks such as Flamingo \cite{alayrac2022flamingo} (cross-modal attention gates) and LLaMA-Adapter \cite{zhang2023llama} (learnable visual projectors) preserve frozen language models while aligning visual features (e.g., CLIP \cite{radford2021learning,ramesh2022hierarchical}) with text embeddings through lightweight modules. End-to-end approaches like LLaVA \cite{liu2023visual} demonstrate that joint vision-language instruction training can unlock deeper multimodal reasoning.

\section{LatticeWorld Framework}\label{sec:problem_setting}
\begin{figure*}[!t]
	\centering
	\includegraphics[width=\linewidth]{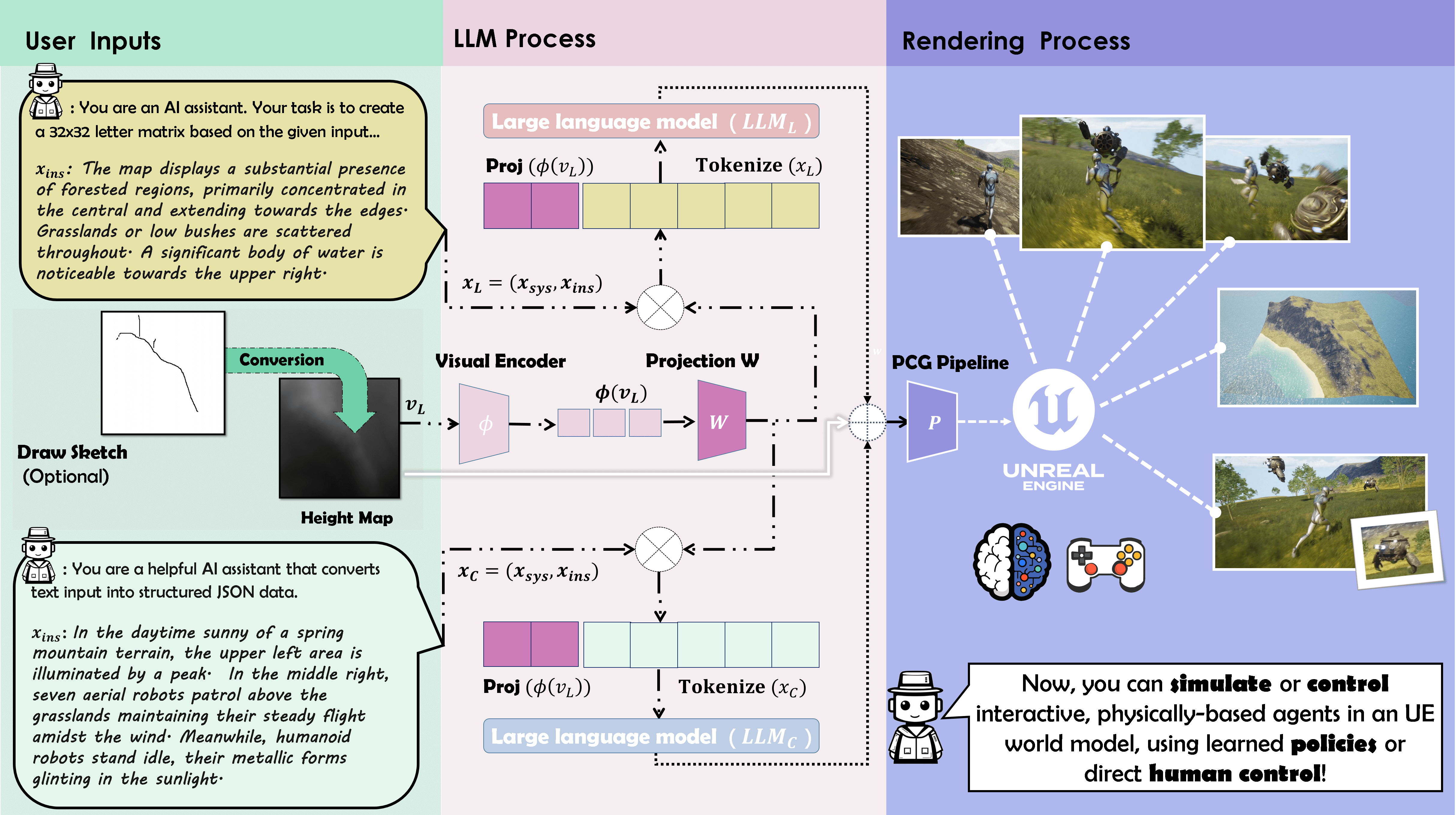}
	\caption{Technical framework of LatticeWorld
	}
	\label{fig:overview}
\end{figure*}

LatticeWorld aims to generate customizable 3D virtual world controlled by users' multimodal instructions, including textual and visual descriptions of the desired environment. To achieve this goal, LatticeWorld encompasses a multimodal scene layout generator, an environmental configuration generator, and a rendering pipeline, as illustrated in \Cref{fig:overview}.

Specifically, LatticeWorld takes in 
the multimodal layout control instructions, comprising a 
textual description $x_\mathrm{L}$ and a vision condition $v_\mathrm{L}$ that contains 3D spatial information, and the text description $x_\mathrm{C}$ of environmental configurations as user-specified inputs. In our work, we refer to the vision condition $v_\mathrm{L}$ as height maps, which can be converted from hand-drawn sketches of the terrain using specific algorithms. Then, LatticeWorld produces a 3D virtual world $\mathcal{W}$ by first employing two well-tuned foundation models, i.e., $\mathsf{LLM}_\mathrm{L}$ and $\mathsf{LLM}_\mathrm{C}$, for the scene layout and environmental configuration generation respectively, that are aligned with users' inputs, and further conducting 3D rendering based on such generated layout and rendering parameters. More formally, we formulate such a generating process as follows:  
\begin{align}
   \hat{y}_\mathrm{L}  &= \mathsf{LLM}_\mathrm{L} \big(x_\mathrm{L},  \Phi(v_\mathrm{L}) \big) \label{eq:LLML} \\
   \hat{y}_\mathrm{C}  &= \mathsf{LLM}_\mathrm{C} \big(x_\mathrm{C},  \Phi(v_\mathrm{L}),  \hat{y}_\mathrm{L}\big) \label{eq:LLMP}\\
   \mathcal{W} &= \mathsf{Render} \big(\Psi_\mathrm{L}(\hat{y}_\mathrm{L}),\Psi_\mathrm{C}(\hat{y}_\mathrm{C}),v_\mathrm{L}\big), \label{eq:render}
\end{align}
where $\Phi$ is the vision-to-word embedding operator that embeds the vision information into the target language space. Here $\Psi_\mathrm{L}$ is a decoder to interpret the intermediate layout representations generated by $\mathsf{LLM}_\mathrm{L}$ into an engine-readable scene layout in a tensor form, $\hat{y}_\mathrm{L}$ denotes such a generated intermediate layout representation. In addition, $\hat{y}_\mathrm{C}$ is the generated environmental configuration, and $\Psi_\mathrm{C}$ represents a configuration translation process to interpret the generated configurations into engine native properties. Here, $\Psi_\mathrm{C}$ can be implemented either through developing translating scripts or using specialized plugins developed based on software like Houdini. $\mathsf{Render}$ denotes the 3D rendering engine for executing the rendering pipeline. It is worth noting that the vision condition $v_\mathrm{L}$ can be optional and is not always required, thus offering flexibility in the application of LatticeWorld. Eventually, we create a playable virtual world, where the main agent can be controlled to interact with other virtual agents. The main agent's actions are currently controlled through input devices, but this can be easily expanded to AI algorithmic policies by using existing UE plug-ins. Our primary focus is on the methodology for creating virtual worlds, while the engineering aspects can be addressed in future development.
Next, we will delve deeper into the three procedures depicted by \eqref{eq:LLML}, \eqref{eq:LLMP}, and \eqref{eq:render} in Sections \ref{sec:language_des}, \ref{sec:pcg-gen}, \ref{sec:agent}, and \ref{sec:render}.

\section{Scene Layout Generation} \label{sec:language_des}
In this section,  we aim to train a scene layout generation model with multimodal inputs of linguistic descriptions and visual conditions integrating LLMs by exploiting and developing LLMs' capabilities of spatial comprehension and symbol sequence generation.

Recall that the generation of scene layout is formulated as $\hat{y}_\mathrm{L}  = \mathsf{LLM}_\mathrm{L} (x_\mathrm{L},  \Phi(v_\mathrm{L}))$ in \eqref{eq:LLML}, which has the textual description $x_\mathrm{L}$ and visual condition $v_\mathrm{L}$ as inputs and produces a layout representation in a specific structure. We target to train $\Phi$ and $\mathsf{LLM}_\mathrm{L}$ to obtain such a scene layout generative model.

Following the same input/output data formats for $\mathsf{LLM}_\mathrm{L}$, the training dataset $\mathcal{D}_\mathrm{L}^\mathrm{tr}$ includes the layout representation $y_\mathrm{L}$, which is transformed from its original layout image $L$. We remark that such an intermediate layout representation is crucial for efficient training and can be decoded into an engine-readable layout tensor via $\Psi_\mathrm{L}$ as in \eqref{eq:render}, eventually resulting in a realistic scene through a rendering engine. Next, we provide the details of the layout representation.

\subsection{Sequential Symbolic Representation of Layout}

In our problem, a scene layout is an image in which different types of assets, characterized by distinct colors, are distributed. Vision-language foundation models like Stable Diffusion have demonstrated remarkable capabilities in visual generation tasks, which may have the potential to be employed for producing layout images directly from textual and visual instructions. However, such models may have limitations of uncontrollable layout generation.
We would like to emphasize that our work focuses on developing a generative framework designed to efficiently leverage a variety of foundation models as a base, even those with merely text generation abilities, rather than concentrating on enhancing vision-language foundation models. 

\begin{table}[!t]
	\centering
	
	\caption{Symbolic Representations of Asset Categories}
		{\small	
			\begin{tabular}{|c|c|c|}
				\hline
				\textbf{Symbol} & \textbf{LoveDA Dataset} & \textbf{Wild Dataset} \\
				\hline
				A & Farmlands & \textemdash \\
				\hline
				B & Buildings & Low bushes or grasslands \\
				\hline
				D & Barren lands & --- \\
				\hline
				F & \multicolumn{2}{c|}{Forests} \\
				\hline
				G & Grasslands & \textemdash \\
				\hline
				R & Roads & Rocky lands \\
				\hline
				S & \textemdash & Snow-capped regions \\
				\hline
				W & \multicolumn{2}{c|}{Water (lakes, rivers, etc.)}\\
				\hline
		\end{tabular}}
		\label{tab:dataset_comparison}
	\end{table}

To tackle the above challenges, we propose an effective intermediate representation scheme to encode a layout image, making it easier for an LLM to incorporate the layout's information via fine-tuning. Specifically, we first compress a scene layout image $L$ into a $p\times p$ symbol (letter) matrix (this work simply lets $p=32$), where each letter symbolizes a certain type of asset. A detailed delineation of such symbolic representations for various datasets is presented in \Cref{tab:dataset_comparison}.
Such a symbolic matrix is further converted into a string $y_\mathrm{L}$ that a LLM can efficiently process: 
\begin{center}
    $y_\mathrm{L} =$ ``$\mathtt{s_1^1s_1^2\cdots s_1^p}$ \verb|\n| $\mathtt{s_2^1s_2^2\cdots s_2^p} $\verb|\n| $\cdots$ \verb|\n| $\mathtt{s_p^1s_p^2\cdots s_p^p}$ '',
\end{center}
where \verb|\n| represents a line break and $\mathtt{s_i^j}$ corresponds to the entry in the $i$-th row and $j$-th column of the matrix. This allows for the conversion of the layout into a fixed-length ``layout symbolic language'' and embeds spatial information within a symbol sequence. Thus, our proposed layout representation can be directly applied to LLMs even with merely text generation abilities, exhibiting its generalizability to various foundation models.  

This layout representation can effectively capture intricate information within the layout, such as the positions and sizes of various asset regions as well as their spatial relationships. For instance, two consecutive letters indicate the adjacency of two regions, and each letter's position corresponds to the respective region's location in the layout. In a favorable format that is readily processable by language models, this sequential symbolic representation can adequately exploit LLMs' strong abilities of sequence understanding and reasoning, enabling the model to generate a logical layout from textual and visual descriptions. 

We remark that an annotation procedure for labeling objects in the training dataset is necessary to guarantee a controllable generation of the layout. Our proposed symbolic representation is equivalent to annotating different types of areas in the layout image. Many existing multimodal foundation models, such as LLM-grounded Diffusion \cite{lian2023llm} and LLaVA, rely on bounding boxes with language descriptions to localize objects in the image. However, the assets in a layout can have particularly irregular shapes, making the bounding box annotation method impractical and inaccurate. Our approach enjoys the flexibility in labeling the regions' types and delineating their shapes. Our method can treat each element in the symbolic matrix as a mini-patch, aligning well with the rendering mechanism in trending engines like UE, where the scene and the terrain are constructed in a grid-based manner.

\subsection{Text-to-Layout for Fixed-Height Scene}\label{sec:fix-height}
We now consider the method for generating layout representations of scenes with minimal terrain variations, i.e., fixed-height scenes, as in the LoveDA dataset that will be used for our experiment. In the later subsection, we further demonstrate that by incorporating visual instructions, variable-height scenes can be further generated.
Specifically, with the model $\mathsf{LLM}_\mathrm{L}$ well trained via instruction tuning, once given a layout instruction $x_\mathrm{L}$, the associated symbolic sequence is generated by $$\hat{y}_\mathrm{L} = \mathsf{LLM}_\mathrm{L}(x_\mathrm{L})$$ as in \eqref{eq:LLML} without the input of the visual instruction $v_\mathrm{L}$.

\subsection{Fusing Vision for Variable-Height Scene} \label{sec:visual_instrc}
In the prior subsection, we simplify the scene generation task by fixing height variations and limiting the user instruction to text. To fully explore LLMs' potential in understanding structured 3D spatial information for layout generation, we need to further encode height information into the model.

To address such a challenge, we propose a multimodal method that can incorporate visual instructions such as the height map into the model, e.g., the terrain height variation can be represented by a grayscale image. This multimodal approach imposes stricter constraints on the layout generation, ensuring more realistic and coherent scene compositions (e.g., snow is on mountaintops rather than lakes). In a height map, a pixel's position can represent the coordinate (latitude and longitude) in reality, while its value indicates the height, which thus can depict spatial height variations. This ensures a strict correspondence between a layout and its height variation. 

In CG industrial scene production, height maps are traditionally crafted by artists using professional software such as World Machine. In view of this problem, beyond the direct height map input, our model allows sketch drawing as a visual instruction. Then we develop a translation model that converts sketch drawings into height maps, trained on strictly corresponding pairs of sketches and height maps. Hence, it simplifies the height map creation process, and users only need to provide a simple sketch to create a detailed, high-resolution height map. Inspired by the recent work~\cite{guerin2017interactive}, our approach leverages a Pix2PixHD model, tailored to our specific task, 
to implement a GAN-based sketch-to-heightmap generative model. It utilizes two different colors of lines to control the height map generation: blue strokes indicate low-lying areas, while black strokes represent ridges.

\vspace{5pt}
\noindent\textbf{Visual Information Integration.} Our multimodal approach integrates an LLM with a sophisticated visual module for encoding visual information, as illustrated in \Cref{fig:overview}. Recall in \eqref{eq:LLML}, given a visual instruction $v_\mathrm{L}$, we obtain a symbolic layout as  $$\hat{y}_\mathrm{L}  = \mathsf{LLM}_\mathrm{L} (x_\mathrm{L},  \Phi(v_\mathrm{L})),$$ where $\Phi$ consists of a visual feature encoder and a projection module that maps visual information associated with $v_\mathrm{L}$ into the LLM's language word embedding space.
At the core of our visual module is the CLIP visual encoder (ViT-B/32), denoted as $\phi$, which generates patch-wise visual features $\phi(v_\mathrm{L})$ for each image $v_\mathrm{L}$ by extracting the output of the penultimate transformer layer.

To map visual features into the word embedding space, we employ a lightweight CNN-based projection network. Specifically, we apply this trainable projection network to convert the patch-wise visual features $\phi(v_\mathrm{L})$ into language embedding tokens $\Phi(v_\mathrm{L})$, which have the same dimensionality as the word embedding space in the LLM:
$$\Phi(v_\mathrm{L}) := \mathsf{Proj}\big(\phi(v_\mathrm{L})\big).$$
This projection enables seamless processing of visual features alongside texts in the LLM's transformer layers.

Our embedding scheme, conceptually inspired by LLaVA \cite{liu2023visual}, demonstrates computational efficiency while effectively transforming visual information into a format that aligns with the LLM's architecture, facilitating joint understanding of visual and textual inputs.

\subsection{Model Training} \label{sec:train-model}

We fine-tune the model $\mathsf{LLM}_\mathrm{L}$ for scene generation via constructing the training data $\mathcal{D}_\mathrm{L}^{\mathrm{tr}}$ for fixed-height scene and variable-height scene generation respectively. 

For the fixed-height scene generation task, the training data $\mathcal{D}_\mathrm{L}^{\mathrm{tr}}$ is a set of data pairs $(x_\mathrm{L}, y_\mathrm{L})$. In our work, we utilize GPT-4o as a sophisticated annotator, generating the textual description $x_\mathrm{L}$ that depicts the scene from various perspectives. We design a series of effective prompts incorporated into $x_\mathrm{L}$, which interprets the symbols in $y_\mathrm{L}$ into real-world concepts, e.g., `` W: Water bodies like lakes, rivers, etc.'' Our symbolic layout representation enables direct mapping of contextual relationships from the serialized symbol matrix to spatial relationships. The textual description $x_\mathrm{L}$ consists of a system prompt $x_{\mathrm{L}}^{\mathrm{sys}}$ defining the task and a user instruction $x_{\mathrm{L}}^{\mathrm{ins}}$ describing the scene, i.e., $x_\mathrm{L} = (x_{\mathrm{L}}^{\mathrm{sys}}, x_{\mathrm{L}}^{\mathrm{ins}})$. In particular, the system prompt $x_{\mathrm{L}}^{\mathrm{sys}}$ is in the following format: \emph{``You are an AI assistant. Your task is to create a 32x32 letter matrix based on the given input. Each line of the matrix should end with `\texttt{\symbol{`\\}n}'. The meaning and distribution of letters in the matrix should accurately reflect the content and visual encoding information provided. The letters in the matrix represent different geographical features: S: Snow-capped mountains or snowy areas, R: Rocky areas or land with many rocks,
 ......''} Based on our curated layout dataset $\mathcal{D}_\mathrm{L}^\mathrm{tr}$, we fine-tune the model LLaMA-2-7B using the cross-entropy loss (supervised fine-tuning).

For the variable-height scene generation task, we conduct fine-tuning of the model integrating visual information, using our vision-language scene training datasets $\mathcal{D}_{\mathrm{L}}^{\mathrm{tr}}$ that are constructed from our proposed Wild dataset, with substantial data augmentation and annotations. As the visual information is now introduced into the framework, we construct a more intricate training dataset $\mathcal{D}_{\mathrm{L}}^{\mathrm{tr}}$ than that in \Cref{sec:fix-height}. In particular, we define $\mathcal{D}_{\mathrm{L}}^{\mathrm{tr}}$ as a set of data tuples $(x_\mathrm{L}, y_\mathrm{L}, v_\mathrm{L}, c_v)$, where $x_\mathrm{L}$ and $y_\mathrm{L}$ are the layout description and layout symbolic representation as in \Cref{sec:fix-height}, $v_\mathrm{L}$ is the visual instruction (e.g., height map), and $c_v$ is the caption of $v_\mathrm{L}$ (e.g., terrain description of a height map). Based on the training dataset, we propose a three-stage training scheme for the layout generation with the incorporation of the visual instructions:
\begin{enumerate}[leftmargin=12pt]
	\item \emph{CLIP fine-tuning for terrain understanding}. 
    To enable the model to extract fine-grained height variation features, we fine-tuned CLIP using our annotated data of height maps $v_\mathrm{L}$  (grayscale images encoding terrain elevation) paired with corresponding captions $c_v$, resulting in an effective visual feature extractor for a height map. 
	\item \emph{Continual pre-training for feature alignment}. In this stage, we keep both the visual encoder in the CLIP model and $\mathsf{LLM}_\mathrm{L}$ weights frozen, focusing on continual pre-training of the projection module $\mathsf{Proj}$. We continue with the height map caption data pairs $(v_\mathrm{L}, c_v)$ used in CLIP fine-tuning. 
    The training of $\mathsf{Proj}$ leverages single-turn conversations, where each sample consists of CLIP-extracted visual features $\phi(v_\mathrm{L})$, a task-specific instruction prompt $x_{\mathrm{ins}}^v$, and the corresponding ground-truth image caption $c_v$. 
    Keeping $\mathsf{LLM}_\mathrm{L}$ parameters frozen, we train $\mathsf{Proj}$ to align visual features with language representations by minimizing the discrepancy between model predictions and ground-truth captions $c_v$. Specifically, $\mathsf{Proj}$ projects CLIP features into language word tokens, which are then concatenated with tokenized $x_{\mathrm{ins}}^v$ - a prompt designed to elicit concise terrain height descriptions - so that we have the model prediction as  $\mathsf{LLM}_\mathrm{L}(\mathrm{concatenate}(\mathsf{Proj}(\phi(v_\mathrm{L})),\mathsf{Tokenize}(x_{\mathrm{ins}}^v)))$. This can establish a robust bridge between visual and linguistic representations through the learned $\mathsf{Proj}$.
	\item \emph{End-to-end fine-tuning}. The final stage involves end-to-end training using the data tuples $(x_\mathrm{L}, v_\mathrm{L}, y_\mathrm{L})$.  
    In this phase, we freeze the visual encoder weights in the CLIP model and fine-tune the pre-trained projection module $\mathsf{Proj}$ and the layout foundation model $\mathsf{LLM}_\mathrm{L}$ (LLaMA-2-7B), enabling more targeted adaptation of both components while preserving the robust visual understanding.

\end{enumerate}
With the above three training stages, we eventually obtain a layout generator with textual and visual instruction inputs.

\section{Environmental Configuration Generation}\label{sec:pcg-gen}
Inspired by the industrial PCG production pipeline, we next concentrate on setting the environmental configurations, a crucial step following the establishment of scene layouts. Since LatticeWorld aims to construct a dynamic environment with interactive agents, our environmental configuration involves two critical aspects:
\begin{itemize}
	\item \emph{Scene attributes}: visual characteristics and spatial arrangement of assets in the scene;
	\item \emph{Agent parameters}: comprehensive agent settings, including their categories, appearance, spatial positions, and behaviors.
\end{itemize}

However, the high complexity and diversity of the environmental configuration incur significant challenges for directly setting them manually, particularly for non-professionals. There may be up to thousands of parameter combinations controlling specific visual effects for a single virtual environment. Moreover, users must not only comprehend the meaning of each parameter but also possess artistic sensibility to achieve optimal effects via manual editing. In our framework, we simplify such a process via an environmental configuration generative model with a paragraph of natural language description and an image as inputs, as we have described in \eqref{eq:LLMP}, i.e., 
$$
\hat{y}_\mathrm{C} = \mathsf{LLM}_\mathrm{C} \big(x_\mathrm{C},  \Phi(v_\mathrm{L}),  \hat{y}_\mathrm{L}\big),
$$ 
where $x_\mathrm{C}$ is the textual description of environmental configurations (including both scene attributes and agent parameters), $v_\mathrm{L}$ is the visual condition (e.g., height maps), $\hat{y}_\mathrm{L}$ is the generated symbolic layout representation from the layout LLM, and $\hat{y}_\mathrm{C}$ is the configuration generation. Moreover, $\Phi$ denotes the vision-to-word embedding operator that has been well trained in \Cref{sec:visual_instrc}. Note that $v_\mathrm{L}$ and $\hat{y}_\mathrm{L}$ are two crucial components in $\mathsf{LLM}_\mathrm{C}$ as environmental configurations can be constrained by the scene layout and terrain elevation, which will be elaborated in \Cref{sec:data}. For example, an aquatic creature should not appear in mountainous terrain. If we consider a fixed-height environment generation, as discussed in \Cref{sec:fix-height}, then the visual embedding module $\Phi(v_\mathrm{L})$ can be removed such that the generation process reduces to $\hat{y}_\mathrm{C} = \mathsf{LLM}_\mathrm{C} \big(x_\mathrm{C},  \hat{y}_\mathrm{L}\big)$.

\subsection{Scene Attributes}\label{sec:scene_attr}
Due to the vast number of scene attributes, directly modeling and mapping them to textual descriptions is challenging and may lead to conflicting generations. To address the above issues, we model the scene attributes in a hierarchical structure that will facilitate an organized translation of languages to attributes, ultimately driving procedural scene detail control and rendering. At the top level are \emph{coarse attributes}, controlling the global setups of the scene, such as the season, the weather, and others. The bottom level consists of \emph{fine attributes}, which further provide a detailed characterization of the generated scene depending on those coarse attributes.

To systematically implement our hierarchical scene attribute approach, we develop a comprehensive attribute transformation framework that formalizes the relationships between different abstraction levels and guides the generation process. Our data generation framework establishes a mapping from natural language descriptions to environmental configurations through a hierarchical attribute system inspired by industrial PCG workflows:

\vspace{5pt}
\noindent\textbf{Coarse Attribute.} The coarse attribute controls the global and general features of the entire environment, encompassing five global aspects of scenes: \emph{terrain type, season, artistic style, weather conditions, and time of day}. These five aspects are associated with specific values for control. For instance, seasons are represented by ``spring'', ``summer'', ``autumn'', and ``winter''.

\begin{table}[!t]
  \centering
  \caption{Discrete (seasonal \& material) and continuous parameters per asset type}
  \small
  \begin{tabular}{|c|c|c|c|}
    \hline
    \makecell{\textbf{Asset} \\ \textbf{Type}}       &\makecell{\textbf{\# Seasonal} \\~\textbf{Params}} & \makecell{\textbf{\# Material} \\ ~\textbf{Params}} & \makecell{\textbf{Continuous} \\ \textbf{Params}} \\
    \hline
    Grass                    & 3                    & 4                    & D                 \\
    Flower                   & 2                    & 4                    & D, R              \\
    Dead Branch              & 2                    & 4                    & D, R              \\
    Stone                    & 3                    & 4                    & D, R              \\
    Architecture             & 4                    & 3                    & D, R, S           \\
    Road                     & 2                    & 2                    & D                 \\
    Lake                     & 2                    & 2                    & D, H              \\
    Desert                   & 6                    & 3                    & D, R, W           \\
    Forest                   & 18                   & 4                    & D, R           \\
    Crops                    & 4                    & 4                    & D, R           \\
    Snow Mountain            & 3                    & 3                    & D, R, Sl          \\
    \hline
    Height Map              & \multicolumn{2}{c|}{N/A} & HM               \\  
    \hline    
    \textbf{Total Discrete}  & 49                   & 37                   & ---               \\
    \hline
    \multicolumn{4}{c}{\vspace{0.1cm} \makecell{\footnotesize D=Density, R=Rotation, S=Scale, H=Height, W=Wind, \\ \footnotesize Sl=Slope, HM=Height Map (Continuous).}} \\
  \end{tabular}    
  \label{tab:symbol_params}  
\end{table}

\vspace{5pt}
\noindent\textbf{Fine Attribute.} There are many types of fine attributes, including discrete parameters 
(e.g., seasonal and material parameters) 
and continuous parameters (e.g., density $D \in [0,1]$). Continuous parameters may be sampled from specific ranges 
. For example, rotations can be represented as three-dimensional Euler angles (pitch, yaw, roll in $[0^\circ,360^\circ)$). Height maps (HM) represent continuous surface geometry but are discretized for storage (typically with values in $[0, 65535]$) and generated using GAN-based methods rather than direct numerical sampling. Intrinsically, these scenes are controlled by coarse scene attributes, the scene's height map characteristics, and layout distribution constraints while adhering to common sense constraints. They also incorporate partial randomization within PCG rules. \Cref{tab:symbol_params} shows different types of assets and the numbers of values for fine attributes, which indicates that in different seasons, asset types have different contents and different surface materials. 
Another type of fine attributes corresponds to the asset placement, including density, orientation, position, etc. We maintain a certain degree of randomness while having certain rule restrictions on those attributes.

\vspace{5pt}
\noindent\textbf{Coarse-to-Fine Transformation.} To integrate the coarse-to-fine generation concept into the language model, we construct the training data in a hierarchical manner for model fine-tuning, which necessitates a coarse-to-fine transformation approach. (Dataset details are presented in Section \ref{sec:env_data}.) 
We implement a rule-based mapping system where coarse attributes define valid ranges and distributions for fine attributes, following industrial PCG principles of hierarchical control. 
For example, selecting ``winter'' as the season automatically adjusts vegetation density parameters downward, restricts available foliage types, and modifies terrain material parameters to include snow coverage. This hierarchical approach can ensure semantic consistency of generation, reduce parameter conflicts in a space containing hundreds of interrelated variables. It also aligns with the workflow of professional artists, who typically begin with high-level scene design before refining finer details. 
Then we achieve a more manageable generation pipeline that can handle complex environments.

\subsection{Agent Parameters}\label{sec:agent}
As discussed so far, the environments created by our method consist of static objects. Next, we investigate building a dynamic environment via incorporating interactive agents/characters into the generated scene. 
Then the created environment has the potential of being developed into a model training platform for embodied AI. Those agents can interact with the main player (main agent) through various dynamic behaviors, including pursuit and combat, creating a competitive platform for studying multi-agent decision-making. 
Specifically, the agent parameters consist of four aspects, including \emph{agents' categories} (such as the goblin, humanoid robot, robotic dog, ancient warrior, etc), \emph{quantities}, \emph{states} (such as idle, patrolling, swimming, etc), and \emph{spatial positions} (such as upper left, lower left, etc), which would allow us to manage the appearance, actions, and interaction strategies of those agents.  In general, the agent parameters are influenced by various scene attributes. For example, a whale should only live in large water bodies. Thus, given $x_\mathrm{C}$, which contains textual description of agent parameters, as well as the generated scene layout $\hat{y}_\mathrm{L}$ and the visual condition $v_\mathrm{L}$, the corresponding agent parameters for agents' categories, states, and spatial positions are created via $\mathsf{LLM}_\mathrm{C}$.
Our generative method enables users to configure and control complex dynamic agents through simple textual and visual instructions, providing a more intuitive and flexible agent control approach for industrial-level dynamic environment design and embodied policy training. 

\subsection{Model Training}
For training the model $\mathsf{LLM}_\mathrm{C}$, we construct the training dataset $\mathcal{D}_\mathrm{C}^\mathrm{tr}$ using a set of data tuples $(x_\mathrm{C}, \Phi(v_\mathrm{L}), y_\mathrm{L},  y_\mathrm{C})$ where $x_\mathrm{C}$ is comprised of the system task definition $x_\mathrm{C}^\mathrm{sys}$ and the environmental configuration description texts $x_\mathrm{C}^\mathrm{ins}$, $y_\mathrm{C}$ consists of the scene attributes and agent parameters in a JSON format, and $v_\mathrm{L}$ and $y_\mathrm{L}$ are the corresponding height maps and symbolic layout representations respectively. During the construction of the training dataset as in \Cref{sec:data}, we constrain both the state and position sets of different agent types as well as the scene attributes based on captions of layouts and terrain elevation for $v_\mathrm{L}$ and $y_\mathrm{L}$. This ensures that both our training data and the resulting outputs accurately align with real-world conditions. 

We train the model to follow specific instructions depicting the key attributes of the scene and the status of agents. We employ LLaMA-2-7B as the base LLM model for our task. Moreover, we apply the visual embedding operator $\Phi$ that is well trained in \Cref{sec:visual_instrc} for projecting the visual condition into the language word tokens. The model training involves fine-tuning the model using the cross-entropy loss and our curated environmental configuration dataset $\mathcal{D}_\mathrm{C}^\mathrm{tr}$ specifically tailored for this task. Eventually, given a textual description of the coarse scene attributes along with agent parameters as well as the height map and the generated layout symbolic representation, the environmental configurations will be generated via $\mathsf{LLM}_\mathrm{C}$.

\section{Procedural Rendering Pipeline}\label{sec:render}
LatticeWorld is an industry-level world model generation framework that includes a complete rendering process using standard PCG pipelines. This would enable rapid migration to various rendering engines and graphics systems to build industrial-grade scenes. For our task, once the symbolic layout representation $\hat{y}_\mathrm{L}$ and environmental configurations $\hat{y}_\mathrm{C}$ are generated via $\mathsf{LLM}_\mathrm{L}$ and $\mathsf{LLM}_\mathrm{C}$, as shown in \eqref{eq:render}, the rendering process can be represented by 
\begin{align*}
	\mathcal{W} = \mathsf{Render} (\Psi_\mathrm{L}(\hat{y}_\mathrm{L}),\Psi_\mathrm{C}(\hat{y}_\mathrm{C}),v_\mathrm{L}), 	
\end{align*}
where $v_\mathrm{L}$ is the visual instructions (e.g., height map) for layout generation, $\mathcal{W}$ is the generated scene after rendering, and $\mathsf{Render}$ represents a rendering engine, e.g., the UE considered in this work. Here, $\Psi_\mathrm{L}$ and $\Psi_\mathrm{C}$ serve as the visual decoder and configuration translator, respectively. 
Notably, our framework can be adapted to various rendering engines, including both industry-grade engines (such as Unity and UE) and non-commercial/open-source engines (such as Blender and Three.js), by only implementing different versions of $\Psi_\mathrm{L}$ and $\Psi_\mathrm{C}$ tailored to the specific input format of each engine. However, due to the inherent limitations of some engines, particularly non-industrial ones, we have chosen to use UE in our work, leveraging its advanced functionalities.

\vspace{5pt}
\noindent\textbf{Layout for Rendering.} Once the symbolic representation of a layout is generated by $\mathsf{LLM}_\mathrm{L}$, we employ the decoder $\Psi_\mathrm{L}$ to translate such a symbolic layout image into the format that the rendering engine can read. In our work, we design a simple yet effective mapping method to implement such a decoder $\Psi_\mathrm{L}$ in three major steps: 
\begin{enumerate}
	
	\item \emph{Layout binary mask creation}. We convert the $p\times p$ symbolic layout ($p=32$ in this paper) into a $p\times p$ low-resolution image, where each character corresponds to a pre-defined RGB color pixel. 
	Then, we create binary (black and white) masks for each color on the RGB image, showing the presence and absence of certain scene types at each pixel.  
	
	\item \emph{Stretching and edge blending}. We stretch those binary masks for each color to the desired size via nearest-neighbor interpolation. Moreover, for a smooth and natural transition of different types of scenes at their edges, 
	we implemented a noise-based edge blending technique, e.g., Gaussian blur, for edge processing such that the binary masks are converted into grayscale images. 
	
	\item \emph{Engine processing}. The rendering engine will understand the scene layout via reading those smoothed mask images for visualization, each corresponding to a distinct scene type. 
	In regions where multiple scene types overlap, the engine can automatically integrate these overlapping elements via a sophisticated blending algorithm, ensuring a natural visual representation. 
\end{enumerate}

\vspace{5pt}
\noindent\textbf{Environmental Configurations for Rendering.} Scene layouts merely express the distribution of scene assets, while the engine also requires necessary configurations to characterize the environment, the assets, and the dynamic agents. The environmental configurations, including the scene attributes and the agent parameters, are generated via $\mathsf{LLM}_\mathrm{C}$, which are further translated into engine native properties via the translation process $\Psi_\mathrm{C}$.
For instance, our weather system leverages the Niagara Fluids plugin \cite{brackbill1988flip,muller2003particle} to implement various weather effects according to the generated parameters, such as sandstorms in desert scenes and snowfall in mountain scenes. 
In other specific implementations, we incorporated various components from UE such as Volumetric Cloud, Volume Fog, SkyBox, SkyAtmosphere, and others.
The environmental configurations are associated with several types of properties: 
\textbf{(1)} \emph{The density and material types of objects}. 
	The environmental configurations can control, for example, vegetation, grasslands, and rock formations across different regions, determining their distribution, diversity, appearances, and visual characteristics. We implement various rules to map these configurations to engine properties to control the density and materials of different objects. 
    \textbf{(2)} \emph{The placement of buildings}. The placement of buildings requires a more nuanced rule-based approach compared to natural elements like grasslands. This process demands consideration of terrain types, height maps, and directional requirements. To address this challenge, we propose building-aware rules to determine building types, positions, and orientations. For example, to enhance realism, we introduce controlled random variations to building orientations and define maximum and minimum distances between buildings. 
    \textbf{(3)} \emph{The configuration of dynamic agents.} The environmental configurations can characterize models, states, initial orientation and distribution, and quantities of the dynamic agents in the rendering engine.
Eventually, $\Psi_\mathrm{C}$ can be implemented either through developing translating scripts or using specialized plugins developed based on software like Houdini. 

\vspace{5pt}
\noindent\textbf{Visual Instruction for Rendering.} The visual instruction $v_\mathrm{L}$ also serves as input to the rendering engine, typically consisting of either a scene height map or a sketch that is convertible to a height map, both of which encode terrain elevation information.

Finally, the rendering engine creates the scene by combining three inputs: the visual information, the generated symbolic layout, and environmental configurations derived through an automated pipeline.

\section{Dataset Construction}\label{sec:data}
Our framework builds upon a well-curated multimodal dataset, which represents another core contribution of this work. To the best of our knowledge, existing multimodal methods for scene generation are incompatible with our proposed framework, resulting in the challenge of lacking data for training our model. For example, methods associated with the Blender engine rely on LLM-generated Blender code and manually crafted rules~\cite{sun20233d,zhou2024scenex,hu2024scenecraft,hu2024scenecraft}. 
To address such a challenge, our work contributes to creating new multimodal datasets tailored for our proposed framework.

\begin{figure}[!t]
\centering
\includegraphics[width=0.7\linewidth]{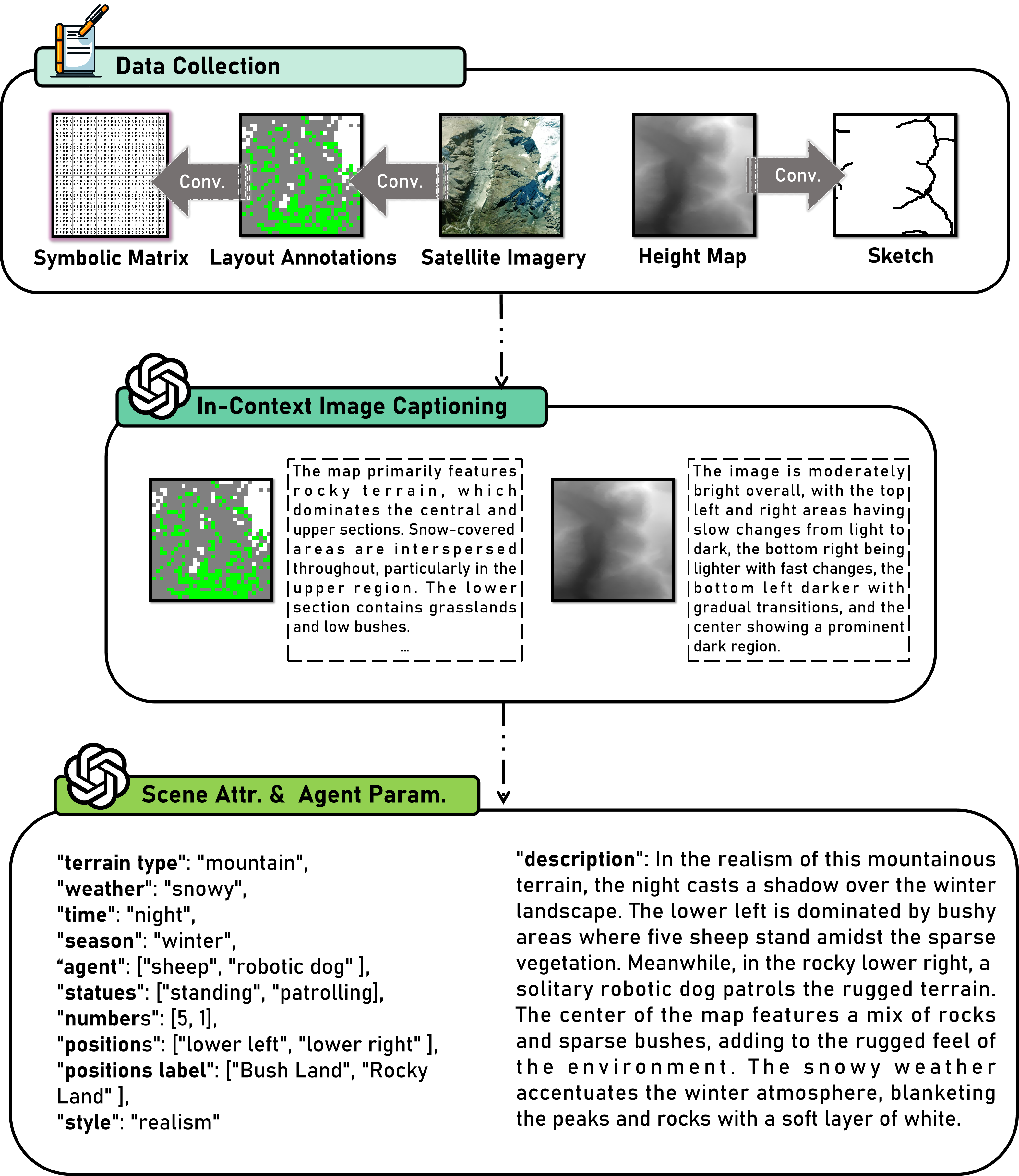}
\caption{Illustration of the dataset construction process.
}
\label{fig:img2layout}
\end{figure}

\subsection{Layout Dataset}
Following our rendering pipeline, to comply with industrial standardization requirements, we transform two raw datasets, the \emph{LoveDA } dataset \cite{junjue_wang_2021_5706578,lovada_data} and our proprietary \emph{Wild} dataset, into multifaceted layout data, including sketches, layout semantic segmentation, and other aspects as detailed below, which are crucial for training and inference. 

LoveDA dataset is an open-source semantic segmentation dataset that contains $5,987$ sensing high spatial resolution (HSR) images. 
Due to the flat terrain in each image, all elevation values for each height map are assigned to $0$. For the Wild dataset, we collected 1,095 high-resolution wilderness scenes from Google Earth Platform. Each image has $2,048\times2,048$ pixels, covering a $5.4$ km$^2$ area with an average pixel resolution of $2.53$ meters. We process those images and their corresponding DEM data by dividing them into $512\times512$-pixel sub-images and creating sketches (by rainfall accumulation algorithms \cite{guerin2017interactive}), height maps (by simulated erosion algorithms), and image semantic segmentation.

Building upon this data, we conduct additional data processing and enhancement in the following steps:
\begin{enumerate}
	\item \emph{Conversion of layouts to symbolic matrices}. We transform the layout semantic segmentation images into $32\times 32$ symbolic matrices illustrated in \Cref{fig:img2layout} . This down-sampling process creates a compact semantic visual-to-text mapping for training $\mathsf{LLM}_\mathrm{L}$. Such a mapping from assets to symbols is detailed in \Cref{tab:dataset_comparison}.
	
	\item \emph{Data augmentation}. To mitigate the risk of overfitting on the dataset and improve the model's robustness, we conduct data augmentation such as rotating images. Then, following the captioning step below, we annotate the same image multiple times from different angles, resulting in multiple descriptions corresponding to a single image and significantly expanding the datasets.
	
	\item \emph{Captioning}. We leveraged the powerful GPT-4o for data annotation as illustrated in \Cref{fig:img2layout}. For the accuracy and consistency of our data annotation process, we developed a sophisticated prompt engineering approach for captioning via GPT-4o based on in-context learning. Particularly, each prompt consists of two major components: \textbf{(1)} \emph{The color-to-scene mapping prompt} that establishes the correspondence between colors and various types of assets; \textbf{(2)} \emph{A layout contextual guidance prompt} that provides specific instructions for describing positions, maintaining conciseness, and preserving adjacency relationships between different types of assets. The annotator, namely GPT-4o, is guided to provide effective spatial relationship and distribution descriptions. For the height map, we also utilize GPT-4o to generate descriptions of elevation changes and their directions.
\end{enumerate}

Finally, we construct such two new datasets for training models following the above procedures as in \Cref{fig:img2layout}. The LoveDA dataset contains $8,236$ data instances derived from the original datasets $2,059$ suburban images. Similarly, the Wild dataset expanded to $24,380$ data instances. Both datasets share a similar structure, with each sample containing layout semantic segmentation, captions, and symbolic matrices. The Wild dataset additionally includes sketches and height maps for each sample, providing more comprehensive geospatial information.

\subsection{Environmental Configuration Dataset}\label{sec:env_data}
We propose a hierarchical framework for constructing environmental configuration and its language description, following industry standards. Then, we build the configuration-description datasets for LoveDA and Wild as follows:

\begin{enumerate}

     \item \emph{Height map and layout captioning}. We use GPT-4o to generate detailed descriptions of layout images and height maps, providing comprehensive textual representations of the environment's visual and spatial characteristics. 
        
    \item \emph{Environmental configuration generation}.   
    Motivated by the UE 5's procedural generation pipeline, we utilize random sampling and structured prompt engineering for GPT4-o to construct a JSON-formatted environmental configuration dataset containing scene attributes and agent parameters.
    We note that these JSON-formatted configurations will be accurately converted to engine-native properties via the transformation function $\Psi_\mathrm{C}$ (\cref{sec:render}) in the rendering process.
    We include two types of configurations: \textbf{(1)} \emph{Context-independent configurations}, which refer to random sampling results that remain applicable across all scene types in our datasets. These include scene attributes (e.g., time of day) and agent parameters (e.g., types). For these configurations, we employ systematic random sampling strategies to maximize coverage of the attribute space while preserving realistic distributions.
    \textbf{(2)} \emph{Context-dependent configurations}, which require common-sense reasoning. We exploit GPT-4o's inherent reasoning capabilities by using structured prompts that integrate previously sampled configurations with height map and layout captioning, enabling contextual analysis for inferring remaining configurations.

    \item \emph{Description generation with GPT-4o}. We employ a rule-based prompt method that integrates the generated context-independent and context-dependent configurations with captioning for height map and layout to guide GPT-4o's inference process and generate comprehensive textual descriptions.

\end{enumerate}

As our future work, we will enrich our datasets with more diverse descriptions, enabling our model to support a wider range of descriptive styles, including more colloquial and conversational language.

\section{Experiment}
In this section, we demonstrate the generation process based on our framework and evaluate its performance via comparison with prior works.

\begin{table*}[!t]
\caption{Comparison of layout generation capabilities on LovaDA and Wild datasets.}
\centering
\newcommand{\adj}[1]{\raisebox{-2pt}[\height][\depth]{#1}}
\resizebox{\textwidth}{!}{%
	\begin{tabular}{ >{\centering\arraybackslash}m{1.9cm}|>{\centering\arraybackslash}m{3.5cm} >{\centering\arraybackslash}m{3.5cm} >{\centering\arraybackslash}m{3.5cm} >{\centering\arraybackslash}m{3.5cm}  }
		
		\hline
		\multicolumn{5}{m{18cm}}{\centering  \textbf{Fixed-Height Layout Generation with Textual Instructions}} 
        \\
        \hline
		\textbf{Instruction} & \multicolumn{4}{m{15.6cm}}{ \vspace{0.05cm}  \textbf{Layout Instruction}: The map displays a mix of land cover types with farmlands occupying the central area, interspersed with buildings. Roads traverse the farmlands and a few buildings are clustered near the center. Bodies of water are mainly on the right, with one extending towards the upper section. Forested regions are on the left, while barren land is scattered throughout, predominantly on the right.} \\  
		\hline
        \vspace{0.05cm} \textbf{Legend} \vspace{0.05cm} & \multicolumn{4}{m{15.6cm}}{%
        \vspace{0.05cm} \centering
        \fcolorbox{white}{green!30}{\phantom{X}} Grassland \;
        \fcolorbox{white}{green}{\phantom{X}} Forest \;
        \fcolorbox{white}[HTML]{FFC280}{\phantom{X}} Farmland \;
        \fcolorbox{white}{blue}{\phantom{X}} Water \;
        \fcolorbox{white}{yellow}{\phantom{X}} Road \;
        \fcolorbox{white}{red}{\phantom{X}} Building \;
        \fcolorbox{white}[HTML]{9D7FB7}{\phantom{X}} Barren
        \vspace{0.05cm}} \\
		\hline
		\multirow{2}*{\textbf{\shortstack[c]{Generated \\ Layout}}}  &\adj{
			\includegraphics[width=\linewidth]{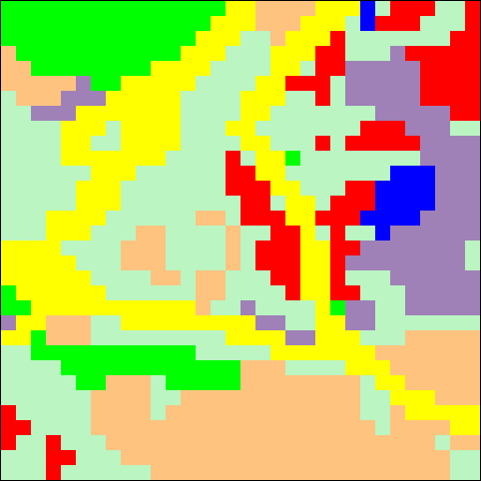}}
		&\adj{\hspace{0.2cm}\includegraphics[width=\linewidth]{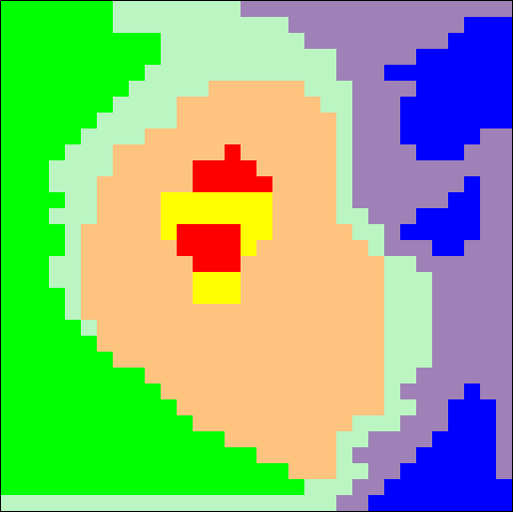}}   &\adj{\hspace{0.35cm}\includegraphics[width=\linewidth]{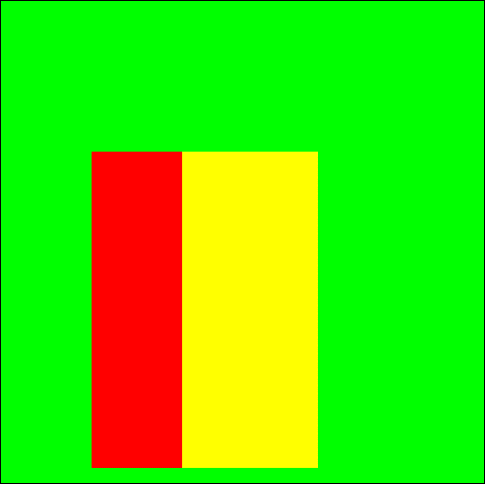}}&\hspace{-0.6cm}    \adj{\includegraphics[width=\linewidth]{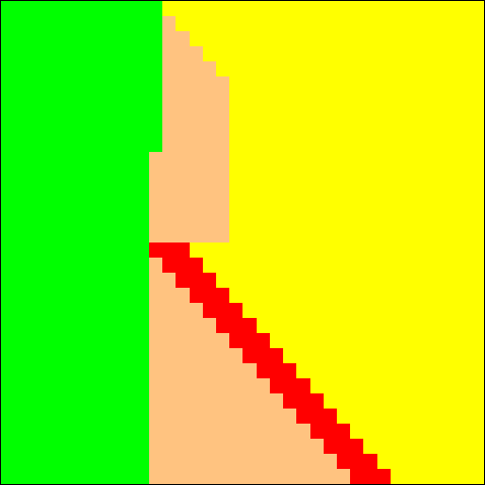}}\\
		& LatticeWorld    &\hspace{0.2cm} Claude 3.7 Sonnet & \hspace{0.35cm} GPT-4o& \hspace{-0.6cm}  DeepSeek-R1\\
		\hline
		\noalign{\vskip 25pt}
		\hline
		\multicolumn{5}{m{18cm}}{\centering  \textbf{Variable-Height Layout Generation with Multimodal Instructions}} \\
		\hline
		\multirow{3}{*}[-6ex]{\textbf{\shortstack[c]{Instruction}}} & \multicolumn{4}{m{15.6cm}}{\vspace{0.05cm} \textbf{Layout Instruction}: The map shows a large area of snow-capped terrain dominating the right side, extending from the upper to the lower regions. Surrounding this, especially concentrated on the left from upper to lower, are regions indicative of low vegetation or grasslands, with patches interspersed throughout the central and right areas, creating a mixed landscape.} \\
		& \multicolumn{4}{m{15.6cm}}{\hspace{4.6cm}\adj{\includegraphics[width=0.12\linewidth]{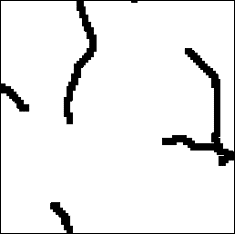}} \qquad \qquad\qquad \adj{\includegraphics[width=0.12\linewidth]{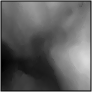}}} \\
		& \multicolumn{4}{m{15.35cm}}{\hspace{4.6cm}~  Sketch \qquad \qquad \qquad \quad    Height Map} \\
		\hline		
        \vspace{0.05cm}\textbf{Legend} \vspace{0.05cm} & \multicolumn{4}{m{15.6cm}}{%
        \vspace{0.05cm} \centering
        \fcolorbox{white}{green}{\phantom{X}} Grassland \;
        \fcolorbox{white}{black}{\phantom{X}} Forest \;
        \fcolorbox{white}[HTML]{808080}{\phantom{X}} Rocky \;
        \fcolorbox{white}[HTML]{0000ff}{\phantom{X}} Water \;
        \fcolorbox{black}[HTML]{ffffff}{\phantom{X}} Snow
        \vspace{0.05cm}} \\
		\hline
		\multirow{2}*{\textbf{\shortstack[c]{Generated \\ Layout}}} &\adj{
			\includegraphics[width=\linewidth]{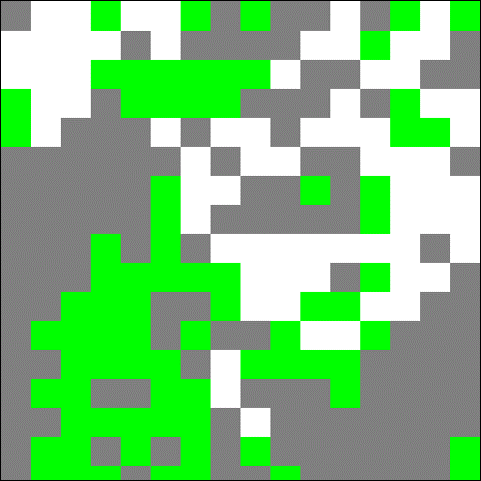}}
		&\adj{\hspace{0.2cm}\includegraphics[width=\linewidth]{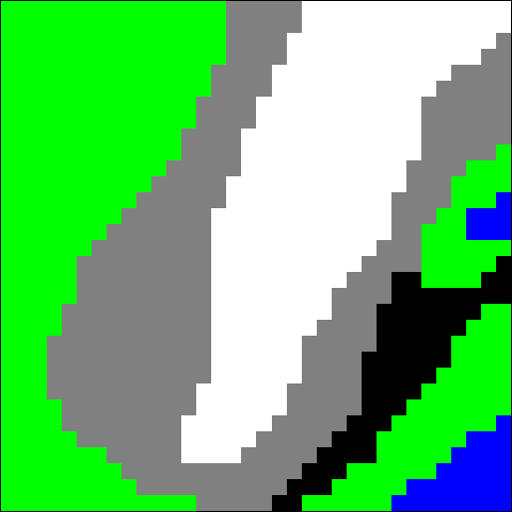}}   &\adj{\hspace{0.35cm}\includegraphics[width=\linewidth]{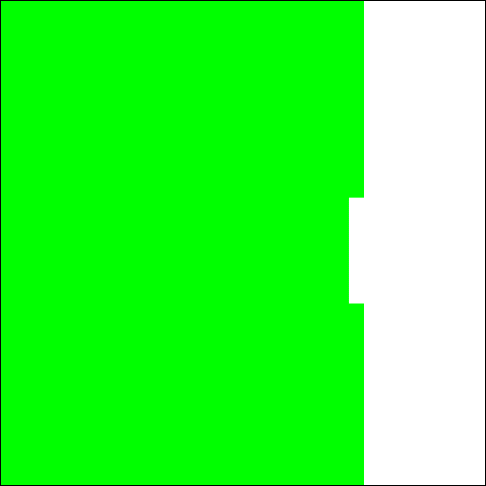}}&\hspace{-0.6cm}   \adj{\includegraphics[width=\linewidth]{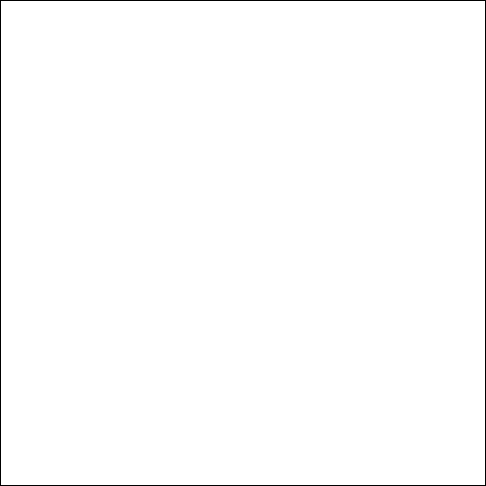}}\\
		& LatticeWorld    &\hspace{0.2cm}Claude 3.7 Sonnet & \hspace{0.35cm} GPT-4o&  \hspace{-0.6cm} Qwen2-VL-Max\\
		\hline
	\end{tabular}%
}
	\label{fig:layout-gen-compare}
    \vspace{0.3cm}
\end{table*}

\begin{table*}[!t]
	\vspace{0.7cm}
	\caption{Demonstration of generated scenes under different multimodal layout instructions.}\label{tab:instruct2scene}
\centering
        \newcommand{\imgH}{2.8cm}
	\newcommand{\adj}[1]{\raisebox{-2pt}[\height][\depth]{#1}}
    \resizebox{\textwidth}{!}{
	\begin{tabular}{ >{\centering\arraybackslash}m{1.3cm}|>{\centering\arraybackslash}m{2.25cm}>{\centering\arraybackslash}m{3.95cm}>{\centering\arraybackslash}m{3.95cm}>{\centering\arraybackslash}m{3.95cm}  }
		\hline
		\multirow{3}{*}[-1.1cm]{\textbf{Demo 1}} & \multicolumn{4}{m{15.2cm}}{ \vspace{0.05cm} \footnotesize \textbf{Layout Instruction}: The map displays a mix of land cover with buildings concentrated towards the center and edges of the upper regions. Roads traverse the map horizontally and vertically, intersecting near the left edge of the central area. Water bodies are scattered throughout, with a more prominent presence in the lower and left areas. The surrounding areas are predominantly covered in farmland, with patches of 
			grassland in the lower left section.} \\[5pt]
		&\hspace{-0.2cm}\adj{
			\includegraphics[width=\linewidth]{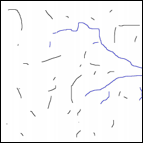}}
		&\adj{\includegraphics[width=\linewidth,height=\imgH,keepaspectratio]{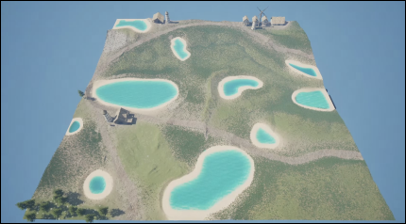}}   &\adj{\includegraphics[width=\linewidth,height=\imgH,keepaspectratio]{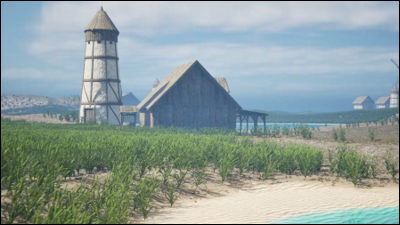}}&\hspace{-0.21cm}   \adj{\includegraphics[width=\linewidth,height=\imgH,keepaspectratio]{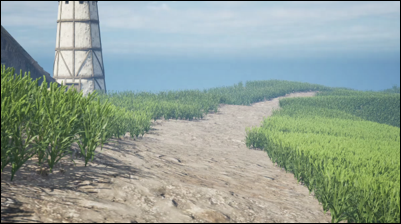}}\\
		
		& {\footnotesize Sketch}    & {\footnotesize Overview} & {\footnotesize Local Scene (i)}&  {\footnotesize  Local Scene (ii)}\\
		\hline
		\hline	
		\multirow{3}{*}[-1.1cm]{\textbf{Demo 2}}  & \multicolumn{4}{m{15.2cm}}{\vspace{0.05cm} \footnotesize \textbf{Layout Instruction}:  The map shows a large area of low bushes and grasslands covering the entire left side and extending to the upper right. Snow-capped mountains or places covered in snow are predominantly located
			in the lower right, with some patches in the upper left. } \\[5pt]
		&\hspace{-0.2cm}\adj{
			\includegraphics[width=\linewidth]{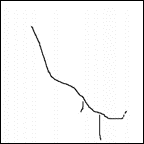}}
		&\adj{\includegraphics[width=\linewidth,height=\imgH,keepaspectratio]{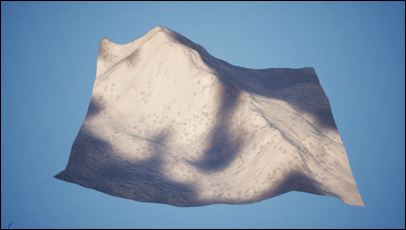}}   &\adj{\includegraphics[width=\linewidth,height=\imgH,keepaspectratio]{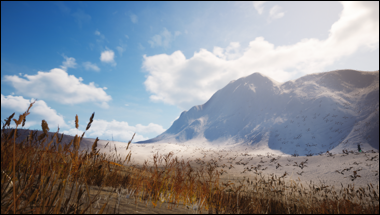}}&\hspace{-0.2cm}   \adj{\includegraphics[width=\linewidth,height=\imgH,keepaspectratio]{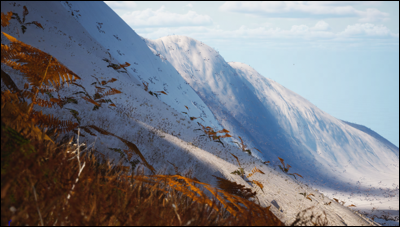}}\\
		& {\footnotesize Sketch}    & {\footnotesize Overview} & {\footnotesize Local Scene (i)}&  {\footnotesize  Local Scene (ii)}\\
		\hline
	\end{tabular}
	}
\vspace{1.2cm}
\caption{Generated scenes under different environmental configuration instructions.}
\centering
\resizebox{\textwidth}{!}{
\begin{tabular}{ >{\centering\arraybackslash}m{1.3cm}|>{\centering\arraybackslash}m{3.0cm}>{\centering\arraybackslash}m{3.6cm}>{\centering\arraybackslash}m{3.6cm}>{\centering\arraybackslash}m{3.6cm}  }
\hline
\textbf{Layout} & \multicolumn{4}{m{15.4cm}}{\vspace{0.05cm} \footnotesize \textbf{Layout Instruction}: The map displays a substantial presence of forested regions, primarily concentrated in the central area and extending towards the edges. Grasslands or low bushes are scattered throughout. A significant body of water is noticeable towards the lower left.} \\[3pt]
\hline
\hline
\multirow{3}{*}[-1.1cm]{\textbf{Demo 1}} & \multicolumn{4}{m{15.4cm}}{\vspace{0.05cm} \footnotesize \textbf{Environmental Configuration Instruction}: In a cartoon-style landscape, the terrain is predominantly mountainous with a sunny autumn day. In the upper left, two eagles are patrolling the skies, their sharp eyes scanning the forested regions below. Meanwhile, in the middle left, an ancient warrior stands idle, perhaps guarding the path through the forest. The central area is lush with forest extending towards the edges, providing a rich tapestry of greens against the backdrop of mountains. To the lower left, a significant body of water glistens under the sun, offering a serene contrast to the rugged peaks. } \\[3pt]
 &\hspace{-0.1cm}\adj{\includegraphics[height=2cm]{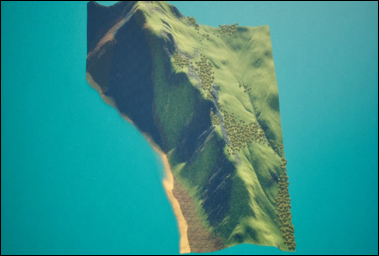}}
  &\adj{\includegraphics[height=2cm]{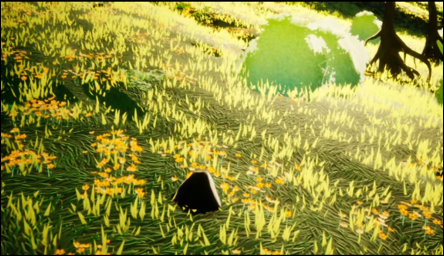}}   
  &\adj{\includegraphics[height=2cm]{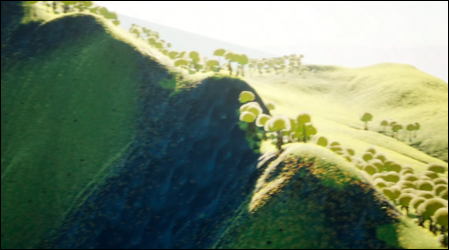}}
  &\hspace{-0.2cm}\adj{\includegraphics[height=2cm]{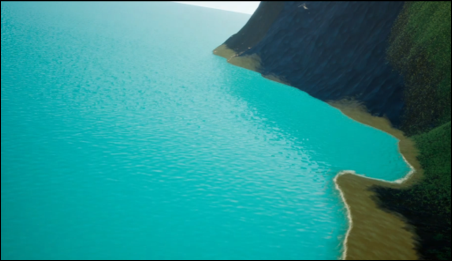}}\\
    &{\footnotesize Overview} & {\footnotesize Local Scene (i)}&  {\footnotesize  Local Scene (ii)}&  {\footnotesize  Local Scene (iii)}\\
\hline
\hline
\multirow{3}{*}[-1cm]{\textbf{Demo 2}} & \multicolumn{4}{m{15.4cm}}{\vspace{0.05cm} \footnotesize \textbf{Environmental Configuration Instruction}: In the upper left of the mountainous terrain, amidst the rain and mist of an autumn afternoon, three ancient warriors patrol the rugged landscape. Their vigilant presence contrasts with the serene middle left, where five sheep are grazing peacefully. The scene captures the realism of a world where ancient and pastoral life coexist, surrounded by the dense forests and the significant body of water noticeable towards the lower left. } \\[3pt]
 &\hspace{-0.1cm}\adj{\includegraphics[height=2cm]{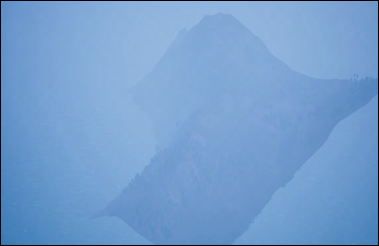}}
  &\adj{\includegraphics[height=2cm]{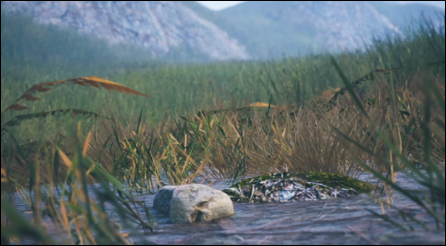}}   
  &\adj{\includegraphics[height=2cm]{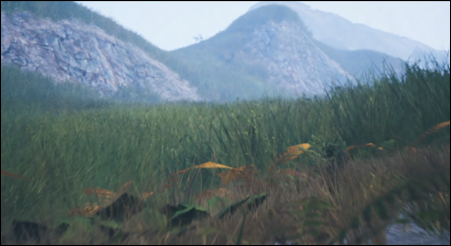}}
  &\hspace{-0.2cm}\adj{\includegraphics[height=2cm]{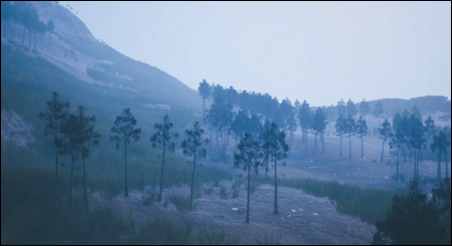}}\\ 
    &{\footnotesize Overview} & {\footnotesize Local Scene (i)}&  {\footnotesize  Local Scene (ii)}&  {\footnotesize  Local Scene (iii)}\\
\hline
\end{tabular}
\label{tab:parameter2scene}
}
\vspace{0.7cm}
\end{table*}

\subsection{Implementation Details}

In the experiments, we train the aforementioned models on our datasets (splitting them into training and test sets) under the following setups: \textbf{(1)} \emph{Fixed-height layout generation with textual inputs}, where we fine-tune LLaMA-2-7B using AdamW optimizer \cite{loshchilov2017decoupled} ($\alpha = 5 \times 10^{-5}$, $\beta_1 = 0.9$, $\beta_2 = 0.999$, $\lambda = 0.001$) for 4 epochs with batch size 32; \textbf{(2)} \emph{Variable-height layout generation with multimodal inputs}, which employs our proposed three-stage training strategy in \Cref{sec:train-model}: CLIP fine-tuning (10 epochs), visual-language feature alignment (12 epochs, $\alpha = 5 \times 10^{-4}$), and end-to-end fine-tuning with frozen CLIP parameters (4 epochs, $\alpha = 5 \times 10^{-5}$, $\beta_1 = 0.9$, $\beta_2 = 0.999$, $\lambda = 0.001$); \textbf{(3)} \emph{Environmental configuration generation}, where we fine-tune LLaMA-2-7B for 5 epochs with batch size 64.
All experiments are conducted on NVIDIA A100 GPUs. Our models are trained using the curated layout datasets as well as the environmental configuration dataset in \Cref{sec:data}. We note that in all experiments, the prompts have been included in the inputs, as introduced in Sections \ref{sec:language_des} and \ref{sec:pcg-gen}. We only highlight user instructions in the results. 

\subsection{Experimental Results}

\begin{table*}[!t]
	\caption{Demonstration of environment generation with dynamic agents.}
    \centering
    \newcommand{\adj}[1]{\raisebox{3pt}{\makebox[\linewidth]{#1}}}
    \resizebox{\textwidth}{!}{
	\begin{tabular}{ >{\arraybackslash}m{5.3cm}| >{\centering\arraybackslash}m{4.9cm} >{\centering\arraybackslash}m{4.9cm}  }
    
		\hline
		\centering \textbf{Instruction} & Local Scene (i) & Local Scene (ii) \\
		\hline 
		{\footnotesize In the lower left of the map, a flock of nine sheep is grazing peacefully on the expansive grassland. Nearby, in the lower right-middle area, two horses are also grazing, enjoying the sunny spring daytime. Above them, in the center of the map, two eagles are patrolling the skies, casting shadows over the landscape. }  &\hspace{0.05cm}\adj{
			\includegraphics[width=\linewidth]{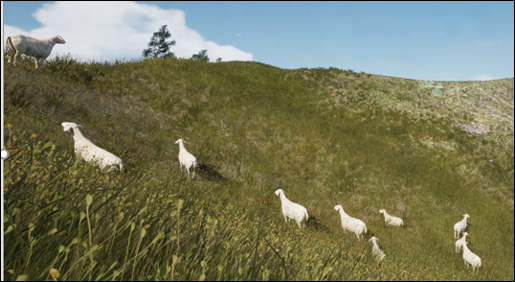}}
		&\hspace{0.05cm}   \adj{\includegraphics[width=\linewidth]{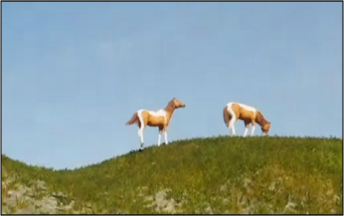}}\\
		\hline 
		{\footnotesize In the daytime sunny of a spring mountain terrain, the upper left area is illuminated by a peak.  In the middle right, seven aerial robots patrol above the grasslands maintaining their steady flight amidst the wind. Meanwhile, humanoid robots stand idle, their metallic forms glinting in the sunlight.} &\hspace{0.05cm}\adj{			\includegraphics[width=\linewidth]{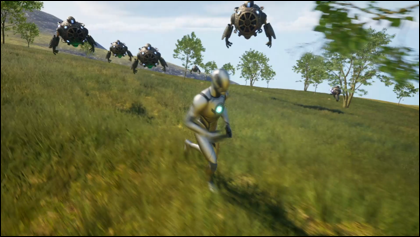}}
		&\hspace{0.05cm}   \adj{\includegraphics[width=\linewidth]{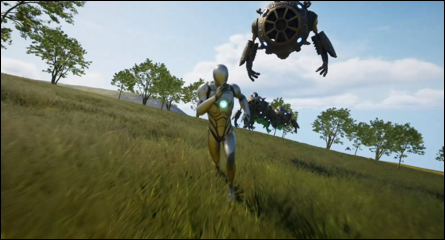}}\\
		\hline 
		{\footnotesize  Under the cover of night, amidst the rainy springtime suburbs, two humanoid robots stand vigilant. The dense cluster of buildings forms the heart of this cyberpunk landscape, their neon lights flickering through the raindrops. One robot is stationed in the center, amidst the towering structures, while the other stands in the lower area, near the buildings. } &\hspace{0.05cm}\adj{
			\includegraphics[width=\linewidth]{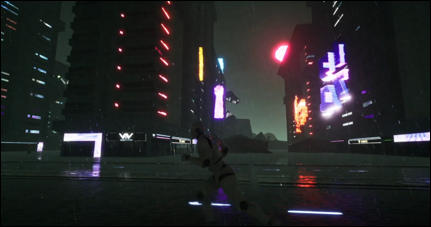}}
		&\hspace{0.05cm}   \adj{\includegraphics[width=\linewidth]{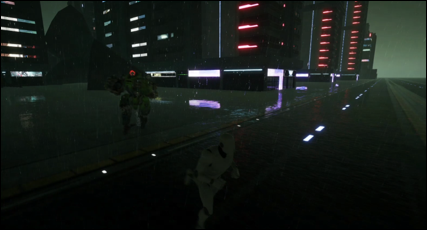}}\\
		\hline
	\end{tabular}
	\label{tab:comparisopn_agent}
    }
\end{table*}

\vspace{5pt}
\noindent\textbf{Environment Generation via LatticeWorld.} 
We evaluate LatticeWorld's text-to-layout generation capabilities against other solutions. Our experiments, as shown in \Cref{fig:layout-gen-compare}, are conducted under two conditions:
\textbf{(1)} \emph{Fixed-Height Condition}. Layout generation using only text descriptions.
\textbf{(2)} \emph{Variable-Height Condition}. Layout generation incorporating visual signals like height maps and sketches.
We compare LatticeWorld with GPT-4o, Claude 3.7 Sonnet\cite{anthropic2024claude}, DeepSeek-R1 \cite{deepseekai2025deepseekr1incentivizingreasoningcapability}, and Qwen2-VL-Max\cite{wang2024qwen2} using identical prompts and instructions. To help other models interpret the height maps, in the experiments for the variable-height condition, we augment the image inputs with captions describing how height information influences the scene. In the results, the $32 \times 32$ symbol matrices are converted into RGB images for visualization.
The results demonstrate LatticeWorld's effectiveness in processing multimodal inputs and generating more accurate layouts across both conditions.

LatticeWorld encodes spatial relationships through a concise $32\times 32$ symbolic matrix. As shown in \Cref{tab:instruct2scene}, we evaluate scene generation with textual and visual inputs. Inspired by the latest industrial workflows, all layouts are rendered in UE 5, with varied terrain while keeping weather and time parameters constant across all experiments.

Moreover, we validate the scene attribute generation ability of LatticeWorld through experiments in \Cref{tab:parameter2scene}, using a fixed layout and various environmental configurations, with different instruction inputs. Our method supports effective generation of diverse natural environments, featuring adjustable scene attributes (e.g., weather and lighting conditions) along with realistic physical property simulation. 

\begin{table*}
	\caption{Qualitative comparisons of generated 3D scenes in wilderness. }\label{tab:comparisopn_3d_wild}
    \centering
	\newcommand{\adj}[1]{\raisebox{-1pt}[\height][\depth]{#1}}
	\setlength{\tabcolsep}{3pt}
	\begin{tabular}{ >{\centering\arraybackslash}m{2.4cm}| >{\centering\arraybackslash}m{3.8cm} >{\centering\arraybackslash}m{3.8cm} >{\centering\arraybackslash}m{3.8cm}  }
		\hline
		\textbf{Methods} & Snow Mountain &River & Forest \\
		\hline
		\textbf{Infinigen} &\hspace{-0.2cm}\adj{
			\includegraphics[width=\linewidth]{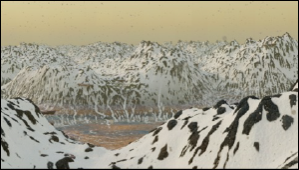}}
		&\adj{\includegraphics[width=\linewidth]{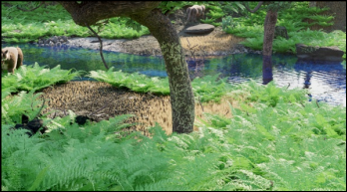}}&\hspace{-0.2cm}   \adj{\includegraphics[width=\linewidth]{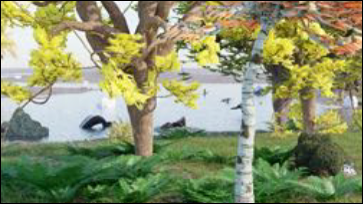}}\\
		\hline
		\textbf{3D-GPT} &\hspace{-0.2cm}\adj{
			\includegraphics[width=\linewidth]{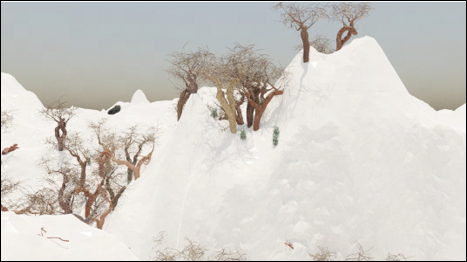}}
		&\adj{\includegraphics[width=\linewidth]{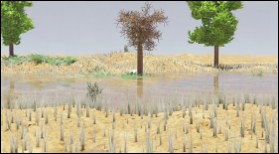}}&\hspace{-0.2cm}   \adj{\includegraphics[width=\linewidth]{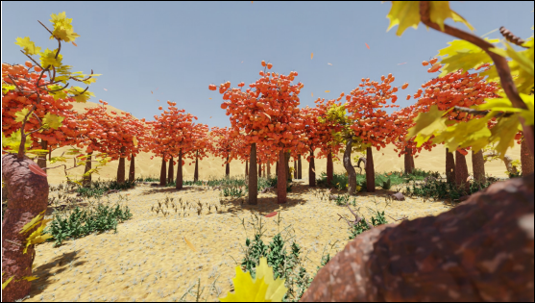}}\\
		\hline
		\textbf{SceneX} &\hspace{-0.2cm}\adj{
			\includegraphics[width=\linewidth]{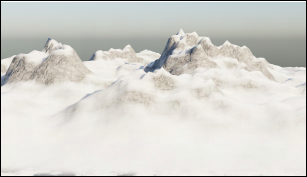}}
		&\adj{\includegraphics[width=\linewidth]{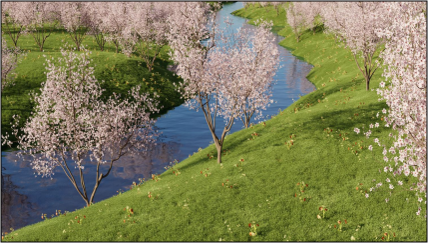}}&\hspace{-0.2cm}   \adj{\includegraphics[width=\linewidth]{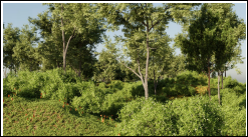}}\\
		\hline
		\textbf{LatticeWorld} &\hspace{-0.2cm}\adj{
			\includegraphics[width=\linewidth]{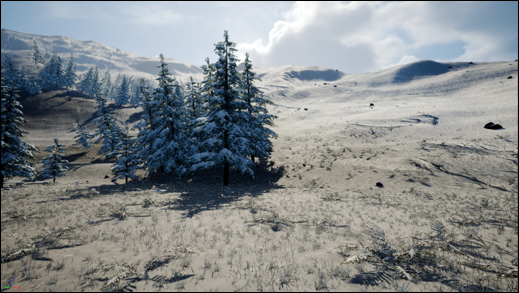}}
		&\adj{\includegraphics[width=\linewidth]{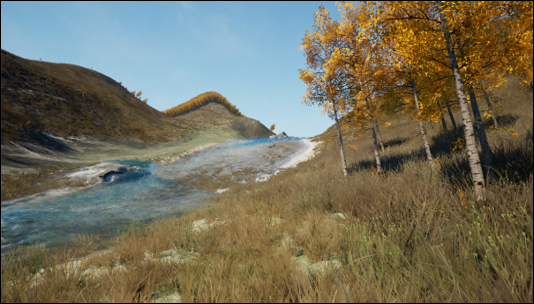}}&\hspace{-0.2cm}   \adj{\includegraphics[width=\linewidth]{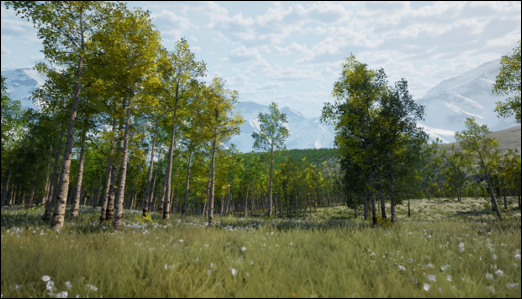}}\\
		\hline
	\end{tabular}
	
    \vspace{0.5cm}

\caption{Qualitative comparisons of generated 3D scenes in rural areas.}\label{tab:comparisopn_3d_rural}
\centering
\begin{tabular}{ >{\centering\arraybackslash}m{2.4cm}|>{\arraybackslash}m{7cm}|>{\centering\arraybackslash}m{4.6cm}  }
\hline
\textbf{Methods} & \centering\textbf{Prompt or Description} &\textbf{Generated Scenes} \\
\hline
\textbf{BlenderGPT} & { Three trees in a row beside a neighborhood.} &\hspace{-0.1cm}\adj{
\includegraphics[width=0.9\linewidth]{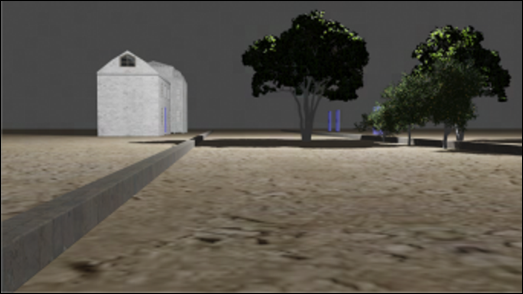}} \\
\hline
\textbf{SceneCraft} & { Three trees in a row beside a neighborhood.} &\hspace{-0.1cm}\adj{
\includegraphics[width=0.9\linewidth]{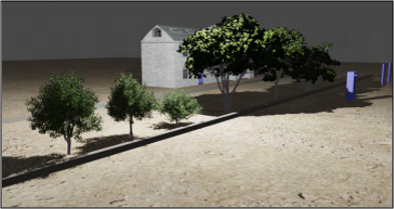}} \\
\hline
\textbf{CPG Landscapes} & { Organized Chambers, Trees, and Rocks.} &\hspace{-0.1cm}\adj{
\includegraphics[width=0.9\linewidth]{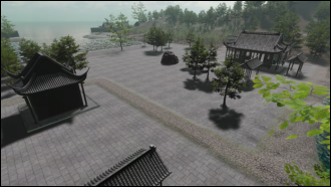}} \\
\hline
\textbf{LatticeWorld} & { The map displays a mix of land cover with buildings concentrated towards the center and upper regions. The surrounding areas are predominantly covered in farmland, with small patches of trees.} &\hspace{-0.1cm}\adj{
	\includegraphics[width=0.9\linewidth]{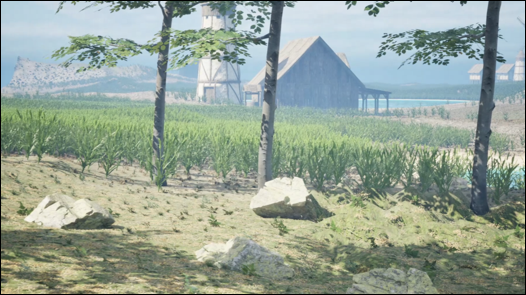}} \\
\hline
\end{tabular}

\end{table*}

\begin{table*}[!t]
	\caption{Qualitative comparisons of generated 3D scenes between LatticeWorld and a human artist.}\label{tab:comparison_3d_human}
    \resizebox{\textwidth}{!}{
	\newcommand{\adj}[1]{\raisebox{-2pt}[\height][\depth]{#1}}
	\begin{tabular}{ >{\centering\arraybackslash}m{2.4cm}|>{\centering\arraybackslash}m{3.35cm}>{\centering\arraybackslash}m{3.35cm}>{\centering\arraybackslash}m{3.35cm}>{\centering\arraybackslash}m{3.35cm}  }
		\hline
        \\[-8pt]
		\multirow{2}{*}[-0.2cm]{\textbf{Instruction}} & \multicolumn{4}{m{14.6cm}}{\small \textbf{Layout Instruction}: The map shows a central road snaking from the lower right to the upper left. Buildings are clustered primarily around this road, with a higher concentration in the center. Forested areas are predominantly on the edges, with a large section in the lower left and smaller patches elsewhere. Water bodies are scattered throughout, and farmlands are situated mostly toward the lower right. Grassland is visible in the whole region, while barren land is interspersed near the center and upper region.}\\[10pt]
		& \multicolumn{4}{m{14.6cm}}{\small \textbf{Environmental Configuration Instruction}: The map is suburbs, season is autumn, realism style in daytime, and today is a sunny day.}\\
		\hline
        \\[-8pt]
		& Overview    &Local Scene (i) &Local Scene (ii) &Local Scene (iii)\\
		
		\textbf{LatticeWorld} &\hspace{-0.2cm}\adj{
			\includegraphics[width=\linewidth]{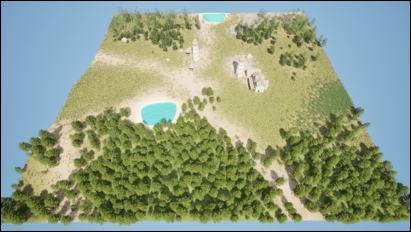}}
		&\adj{\includegraphics[width=\linewidth]{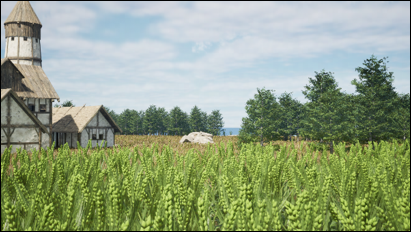}}   &\adj{\includegraphics[width=\linewidth]{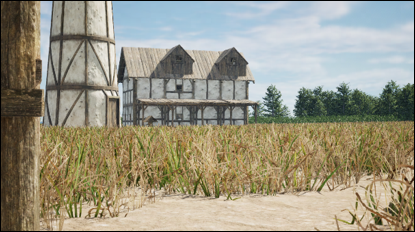}}
		&\adj{\includegraphics[width=\linewidth]{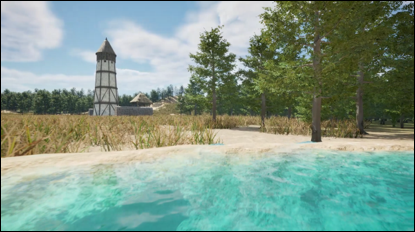}}\\
		
		\textbf{Artist} &\hspace{-0.2cm}\adj{
			\includegraphics[width=\linewidth]{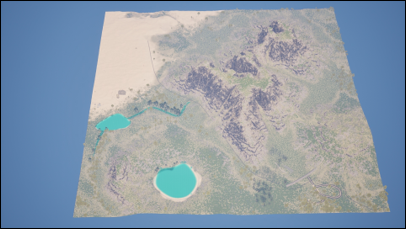}}
		&\adj{\includegraphics[width=\linewidth]{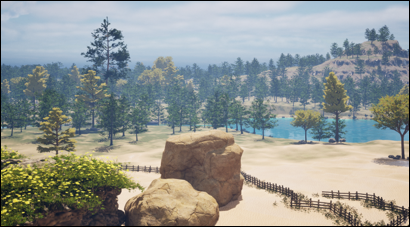}}   &\adj{\includegraphics[width=\linewidth]{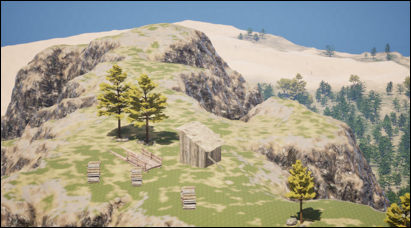}}
		&\adj{\includegraphics[width=\linewidth]{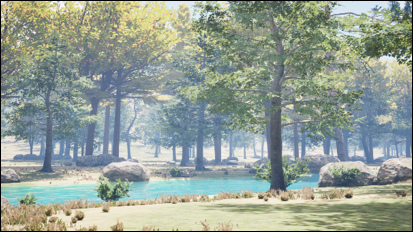}}\\       
		\hline
	\end{tabular}
	\vspace{0.5cm}
    }

\caption{The comparison between LatticeWorld and artists' workload (measured in days).
}\label{fig:workload_comarison}
\centering
\includegraphics[width=0.75\linewidth]{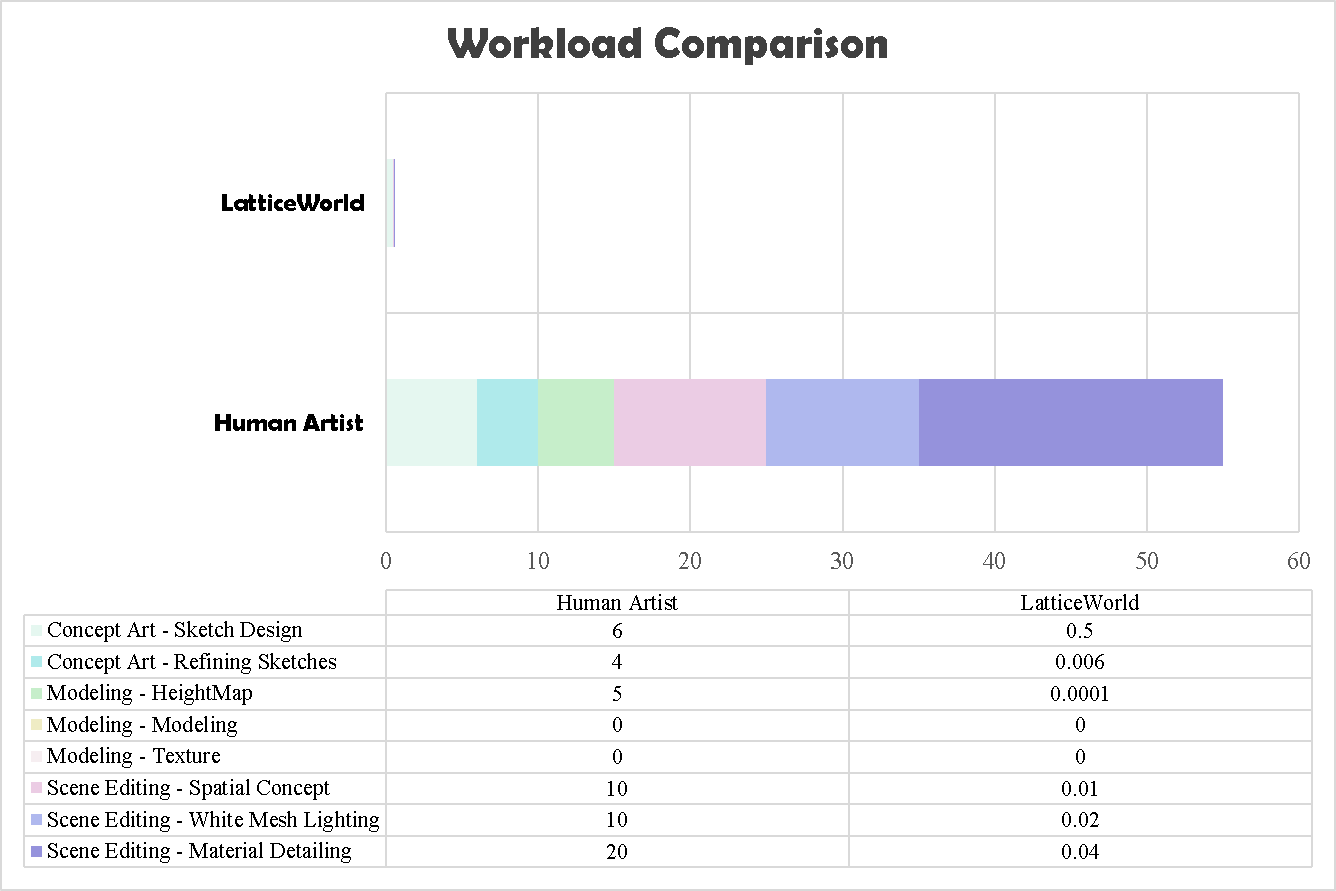}
\vspace{-0.2cm}
\end{table*}

\vspace{5pt}
\noindent\textbf{Dynamic Environment with Interactive Agents.} As shown in \Cref{tab:comparisopn_agent}, we can build multi-agent interaction environments based on LatticeWorld. 
The experiments show that our framework also supports effective configuration of agent parameters (e.g., agent types and quantities). These agents are equipped with environmental perception capabilities and can perform autonomous adversarial behaviors based on predefined rules, e.g., they will automatically pursue and attack the main agent when the main agent enters a specified proximity range. Our framework also allows for future implementation of more sophisticated behavioral policies in agent interactions.
These features position our framework as a potential platform for embodied agent training.

\newpage
\noindent\textbf{Comparison with Prior Works.} 
Since LatticeWorld employs a platform-based paradigm, we compare LatticeWorld with existing platform-based methods, focusing solely on scene generation, since prior methods may lack support for dynamic interactive agents.
As illustrated in \Cref{tab:comparisopn_3d_wild} and \Cref{tab:comparisopn_3d_rural}, we conduct a qualitative comparison using published demonstrations from prior works and show that LatticeWorld achieves better generation quality. 
Due to different methodologies and datasets, we only choose similar scenes for comparison. It is worth noting that LatticeWorld generates large 3D scenes, and we only capture a portion of the scene details.

\vspace{5pt}
\noindent\textbf{Comparison with Industrial Manual Methods.} 
The standard industrial creation of dynamic environments combines PCG with manual artistic work. 
The process consists of three main phases: \textbf{(1)} \emph{the concept art phase} (sketching and refinement), \textbf{(2)} \emph{the modeling phase} (height maps, 3D assets, and UV mapping), and \textbf{(3)} \emph{the scene editing phase} (layout, lighting, and materials).
We compare environments created by professional artists and LatticeWorld using identical layout and parameter instructions. We choose similar scenes containing trees and buildings for comparison, and the results are shown in \Cref{tab:comparison_3d_human}. Workload comparison (\Cref{fig:workload_comarison}) shows that while LatticeWorld uses pre-completed assets for sketching, modeling, and texturing, it significantly improves efficiency in other steps. Specifically, LatticeWorld lowers total production time from $55$ days (manual) to less than $0.6$ days, yielding over a $90\times$ efficiency improvement. This advantage increases when generating multiple environments, as the pre-completion cost is distributed.

\section{Conclusion}

This paper introduces LatticeWorld, a multimodal framework for interactive virtual world generation via LLMs.  LatticeWorld leverages lightweight LLMs alongside the industry-grade rendering engine (e.g., Unreal Engine 5) to generate a dynamic environment with multimodal instructions. Our work on LatticeWorld can be further improved in the following aspects: 
\textbf{(1)} The adversarial agents currently follow a simple policy: they attack the main agent whenever it approaches them. We can implement more diverse policies to create varied adversarial behaviors. \textbf{(2)} Our current framework is limited to controlling a single main player. We can enhance its functionality to control multiple main players. Furthermore, the main agent is currently controlled via input devices. This capability can be expanded to support AI algorithmic policies by using existing UE plug-ins. \textbf{(3)} Main agent's body parts cannot be controlled independently. We can add finer-grained control of specific parts through more sophisticated modeling. \textbf{(4)} Finally, we plan to expand the asset library with more objects and interactive elements to generate more diverse virtual worlds.

\section*{Acknowledgement} 

The authors would like to thank Zixi Liu for his technical assistance and helpful discussions.

\bibliographystyle{plain}
\bibliography{egbib}

\begin{thebibliography}{10}

\bibitem{alayrac2022flamingo}
Jean-Baptiste Alayrac, Jeff Donahue, Pauline Luc, Antoine Miech, Iain Barr,
  Yana Hasson, Karel Lenc, Arthur Mensch, Katherine Millican, Malcolm Reynolds,
  et~al.
\newblock Flamingo: a visual language model for few-shot learning.
\newblock {\em Advances in Neural Information Processing Systems},
  35:23716--23736, 2022.

\bibitem{anthropic2024claude}
AI~Anthropic.
\newblock The claude 3 model family: Opus, sonnet, haiku.
\newblock {\em Claude-3 Model Card}, 2024.

\bibitem{ayubi2020deterministic}
Peyman Ayubi, Saeed Setayeshi, and Amir~Masoud Rahmani.
\newblock Deterministic chaos game: a new fractal based pseudo-random number
  generator and its cryptographic application.
\newblock {\em Journal of Information Security and Applications}, 52:102472,
  2020.

\bibitem{beukman2022procedural}
Michael Beukman, Christopher~W Cleghorn, and Steven James.
\newblock Procedural content generation using neuroevolution and novelty search
  for diverse video game levels.
\newblock In {\em Proceedings of the Genetic and Evolutionary Computation
  Conference}, pages 1028--1037, 2022.

\bibitem{brackbill1988flip}
Jeremiah~U Brackbill, Douglas~B Kothe, and Hans~M Ruppel.
\newblock Flip: a low-dissipation, particle-in-cell method for fluid flow.
\newblock {\em Computer Physics Communications}, 48(1):25--38, 1988.

\bibitem{brown2020language}
Tom Brown, Benjamin Mann, Nick Ryder, Melanie Subbiah, Jared~D Kaplan, Prafulla
  Dhariwal, Arvind Neelakantan, Pranav Shyam, Girish Sastry, Amanda Askell,
  et~al.
\newblock Language models are few-shot learners.
\newblock {\em Advances in neural information processing systems},
  33:1877--1901, 2020.

\bibitem{chen2024text}
Zilong Chen, Feng Wang, Yikai Wang, and Huaping Liu.
\newblock Text-to-3d using gaussian splatting.
\newblock In {\em Proceedings of the IEEE/CVF Conference on Computer Vision and
  Pattern Recognition}, pages 21401--21412, 2024.

\bibitem{deepseekai2025deepseekr1incentivizingreasoningcapability}
DeepSeek-AI, Daya Guo, Dejian Yang, Haowei Zhang, Junxiao Song, Ruoyu Zhang,
  Runxin Xu, Qihao Zhu, Shirong Ma, Peiyi Wang, Xiao Bi, Xiaokang Zhang,
  Xingkai Yu, Yu~Wu, Z.~F. Wu, Zhibin Gou, Zhihong Shao, Zhuoshu Li, Ziyi Gao,
  Aixin Liu, Bing Xue, Bingxuan Wang, Bochao Wu, Bei Feng, Chengda Lu,
  Chenggang Zhao, Chengqi Deng, Chenyu Zhang, Chong Ruan, Damai Dai, Deli Chen,
  Dongjie Ji, Erhang Li, Fangyun Lin, Fucong Dai, Fuli Luo, Guangbo Hao,
  Guanting Chen, Guowei Li, H.~Zhang, Han Bao, Hanwei Xu, Haocheng Wang,
  Honghui Ding, Huajian Xin, Huazuo Gao, Hui Qu, Hui Li, Jianzhong Guo, Jiashi
  Li, Jiawei Wang, Jingchang Chen, Jingyang Yuan, Junjie Qiu, Junlong Li, J.~L.
  Cai, Jiaqi Ni, Jian Liang, Jin Chen, Kai Dong, Kai Hu, Kaige Gao, Kang Guan,
  Kexin Huang, Kuai Yu, Lean Wang, Lecong Zhang, Liang Zhao, Litong Wang, Liyue
  Zhang, Lei Xu, Leyi Xia, Mingchuan Zhang, Minghua Zhang, Minghui Tang, Meng
  Li, Miaojun Wang, Mingming Li, Ning Tian, Panpan Huang, Peng Zhang, Qiancheng
  Wang, Qinyu Chen, Qiushi Du, Ruiqi Ge, Ruisong Zhang, Ruizhe Pan, Runji Wang,
  R.~J. Chen, R.~L. Jin, Ruyi Chen, Shanghao Lu, Shangyan Zhou, Shanhuang Chen,
  Shengfeng Ye, Shiyu Wang, Shuiping Yu, Shunfeng Zhou, Shuting Pan, S.~S. Li,
  Shuang Zhou, Shaoqing Wu, Shengfeng Ye, Tao Yun, Tian Pei, Tianyu Sun,
  T.~Wang, Wangding Zeng, Wanjia Zhao, Wen Liu, Wenfeng Liang, Wenjun Gao,
  Wenqin Yu, Wentao Zhang, W.~L. Xiao, Wei An, Xiaodong Liu, Xiaohan Wang,
  Xiaokang Chen, Xiaotao Nie, Xin Cheng, Xin Liu, Xin Xie, Xingchao Liu, Xinyu
  Yang, Xinyuan Li, Xuecheng Su, Xuheng Lin, X.~Q. Li, Xiangyue Jin, Xiaojin
  Shen, Xiaosha Chen, Xiaowen Sun, Xiaoxiang Wang, Xinnan Song, Xinyi Zhou,
  Xianzu Wang, Xinxia Shan, Y.~K. Li, Y.~Q. Wang, Y.~X. Wei, Yang Zhang,
  Yanhong Xu, Yao Li, Yao Zhao, Yaofeng Sun, Yaohui Wang, Yi~Yu, Yichao Zhang,
  Yifan Shi, Yiliang Xiong, Ying He, Yishi Piao, Yisong Wang, Yixuan Tan,
  Yiyang Ma, Yiyuan Liu, Yongqiang Guo, Yuan Ou, Yuduan Wang, Yue Gong, Yuheng
  Zou, Yujia He, Yunfan Xiong, Yuxiang Luo, Yuxiang You, Yuxuan Liu, Yuyang
  Zhou, Y.~X. Zhu, Yanhong Xu, Yanping Huang, Yaohui Li, Yi~Zheng, Yuchen Zhu,
  Yunxian Ma, Ying Tang, Yukun Zha, Yuting Yan, Z.~Z. Ren, Zehui Ren, Zhangli
  Sha, Zhe Fu, Zhean Xu, Zhenda Xie, Zhengyan Zhang, Zhewen Hao, Zhicheng Ma,
  Zhigang Yan, Zhiyu Wu, Zihui Gu, Zijia Zhu, Zijun Liu, Zilin Li, Ziwei Xie,
  Ziyang Song, Zizheng Pan, Zhen Huang, Zhipeng Xu, Zhongyu Zhang, and Zhen
  Zhang.
\newblock Deepseek-r1: Incentivizing reasoning capability in llms via
  reinforcement learning, 2025.

\bibitem{deng2023citygeninfinitecontrollable3d}
Jie Deng, Wenhao Chai, Jianshu Guo, Qixuan Huang, Wenhao Hu, Jenq-Neng Hwang,
  and Gaoang Wang.
\newblock Citygen: Infinite and controllable 3d city layout generation, 2023.

\bibitem{deng2024citycraft}
Jie Deng, Wenhao Chai, Junsheng Huang, Zhonghan Zhao, Qixuan Huang, Mingyan
  Gao, Jianshu Guo, Shengyu Hao, Wenhao Hu, Jenq-Neng Hwang, et~al.
\newblock Citycraft: A real crafter for 3d city generation.
\newblock {\em arXiv preprint arXiv:2406.04983}, 2024.

\bibitem{earle2022illuminating}
Sam Earle, Justin Snider, Matthew~C Fontaine, Stefanos Nikolaidis, and Julian
  Togelius.
\newblock Illuminating diverse neural cellular automata for level generation.
\newblock In {\em Proceedings of the Genetic and Evolutionary Computation
  Conference}, pages 68--76, 2022.

\bibitem{freiknecht2017survey}
Jonas Freiknecht and Wolfgang Effelsberg.
\newblock A survey on the procedural generation of virtual worlds.
\newblock {\em Multimodal Technologies and Interaction}, 1(4):27, 2017.

\bibitem{scenescape}
Rafail Fridman, Amit Abecasis, Yoni Kasten, and Tali Dekel.
\newblock Scenescape: Text-driven consistent scene generation.
\newblock {\em arXiv preprint arXiv:2302.01133}, 2023.

\bibitem{fridman2024scenescape}
Rafail Fridman, Amit Abecasis, Yoni Kasten, and Tali Dekel.
\newblock Scenescape: Text-driven consistent scene generation.
\newblock {\em Advances in Neural Information Processing Systems}, 36, 2024.

\bibitem{gambi2019automatically}
Alessio Gambi, Marc Mueller, and Gordon Fraser.
\newblock Automatically testing self-driving cars with search-based procedural
  content generation.
\newblock In {\em Proceedings of the 28th ACM SIGSOFT International Symposium
  on Software Testing and Analysis}, pages 318--328, 2019.

\bibitem{gd3kr_BlenderGPT_2023}
gd3kr.
\newblock Blendergpt.
\newblock \url{https://github.com/gd3kr/BlenderGPT}, April 2023.

\bibitem{genie2}
{Genie-2 Team}.
\newblock Genie-2: Advanced interactive virtual environment.
\newblock \url{https://genie2.co/}, 2024.
\newblock Accessed: 2024-12-12.

\bibitem{guerin2017interactive}
{\'E}ric Gu{\'e}rin, Julie Digne, Eric Galin, Adrien Peytavie, Christian Wolf,
  Bedrich Benes, and Beno{\^\i}t Martinez.
\newblock Interactive example-based terrain authoring with conditional
  generative adversarial networks.
\newblock {\em ACM Trans. Graph.}, 36(6):228--1, 2017.

\bibitem{text2room}
Lukas H\"ollein, Ang Cao, Andrew Owens, Justin Johnson, and Matthias
  Nie{\ss}ner.
\newblock Text2room: Extracting textured 3d meshes from 2d text-to-image
  models.
\newblock In {\em Proceedings of the IEEE/CVF International Conference on
  Computer Vision}, pages 7909--7920, 2023.

\bibitem{hu2024scenecraft}
Ziniu Hu, Ahmet Iscen, Aashi Jain, Thomas Kipf, Yisong Yue, David~A Ross,
  Cordelia Schmid, and Alireza Fathi.
\newblock Scenecraft: An llm agent for synthesizing 3d scenes as blender code.
\newblock In {\em Forty-first International Conference on Machine Learning},
  2024.

\bibitem{jiang2023mistral}
Albert~Q Jiang, Alexandre Sablayrolles, Arthur Mensch, Chris Bamford, Devendra
  Singh~Chaplot, Diego de~las Casas, Florian Bressand, Gianna Lengyel,
  Guillaume Lample, Lucile Saulnier, et~al.
\newblock Mistral 7b.
\newblock {\em arXiv e-prints}, pages arXiv--2310, 2023.

\bibitem{kerbl3Dgaussians}
Bernhard Kerbl, Georgios Kopanas, Thomas Leimk{\"u}hler, and George Drettakis.
\newblock 3d gaussian splatting for real-time radiance field rendering.
\newblock {\em ACM Transactions on Graphics (SIGGRAPH)}, 42(4), July 2023.

\bibitem{khalifa2020pcgrl}
Ahmed Khalifa, Philip Bontrager, Sam Earle, and Julian Togelius.
\newblock Pcgrl: Procedural content generation via reinforcement learning.
\newblock In {\em Proceedings of the AAAI Conference on Artificial Intelligence
  and Interactive Digital Entertainment}, volume~16, pages 95--101, 2020.

\bibitem{li2021igibson20objectcentricsimulation}
Chengshu Li, Fei Xia, Roberto Martín-Martín, Michael Lingelbach, Sanjana
  Srivastava, Bokui Shen, Kent Vainio, Cem Gokmen, Gokul Dharan, Tanish Jain,
  Andrey Kurenkov, C.~Karen Liu, Hyowon Gweon, Jiajun Wu, Li~Fei-Fei, and
  Silvio Savarese.
\newblock igibson 2.0: Object-centric simulation for robot learning of everyday
  household tasks, 2021.

\bibitem{li2025dreamscene}
Haoran Li, Haolin Shi, Wenli Zhang, Wenjun Wu, Yong Liao, Lin Wang, Lik-hang
  Lee, and Peng~Yuan Zhou.
\newblock Dreamscene: 3d gaussian-based text-to-3d scene generation via
  formation pattern sampling.
\newblock In {\em European Conference on Computer Vision}, pages 214--230.
  Springer, 2025.

\bibitem{li2024director3d}
Xinyang Li, Zhangyu Lai, Linning Xu, Yansong Qu, Liujuan Cao, Shengchuan Zhang,
  Bo~Dai, and Rongrong Ji.
\newblock Director3d: Real-world camera trajectory and 3d scene generation from
  text.
\newblock {\em arXiv preprint arXiv:2406.17601}, 2024.

\bibitem{lian2023llm}
Long Lian, Boyi Li, Adam Yala, and Trevor Darrell.
\newblock Llm-grounded diffusion: Enhancing prompt understanding of
  text-to-image diffusion models with large language models.
\newblock {\em arXiv preprint arXiv:2305.13655}, 2023.

\bibitem{lin2021softgymbenchmarkingdeepreinforcement}
Xingyu Lin, Yufei Wang, Jake Olkin, and David Held.
\newblock Softgym: Benchmarking deep reinforcement learning for deformable
  object manipulation, 2021.

\bibitem{liu2023visual}
Haotian Liu, Chunyuan Li, Qingyang Wu, and Yong~Jae Lee.
\newblock Visual instruction tuning.
\newblock {\em Advances in neural information processing systems}, 36, 2023.

\bibitem{liu2024controllable}
Jia-Hong Liu, Shao-Kui Zhang, Chuyue Zhang, and Song-Hai Zhang.
\newblock Controllable procedural generation of landscapes.
\newblock In {\em ACM Multimedia 2024}, 2024.

\bibitem{liu2021deep}
Jialin Liu, Sam Snodgrass, Ahmed Khalifa, Sebastian Risi, Georgios~N
  Yannakakis, and Julian Togelius.
\newblock Deep learning for procedural content generation.
\newblock {\em Neural Computing and Applications}, 33(1):19--37, 2021.

\bibitem{loshchilov2017decoupled}
I~Loshchilov.
\newblock Decoupled weight decay regularization.
\newblock {\em arXiv preprint arXiv:1711.05101}, 2017.

\bibitem{lu2024urban}
Fan Lu, Kwan-Yee Lin, Yan Xu, Hongsheng Li, Guang Chen, and Changjun Jiang.
\newblock Urban architect: Steerable 3d urban scene generation with layout
  prior.
\newblock {\em arXiv preprint arXiv:2404.06780}, 2024.

\bibitem{lu2024genexgeneratingexplorableworld}
Taiming Lu, Tianmin Shu, Junfei Xiao, Luoxin Ye, Jiahao Wang, Cheng Peng, Chen
  Wei, Daniel Khashabi, Rama Chellappa, Alan Yuille, and Jieneng Chen.
\newblock Genex: Generating an explorable world, 2024.

\bibitem{llama3_model}
MetaAI.
\newblock Llama 3 model, 2024.
\newblock Accessed: 2024-06-13.

\bibitem{mildenhall2021nerf}
Ben Mildenhall, Pratul~P Srinivasan, Matthew Tancik, Jonathan~T Barron, Ravi
  Ramamoorthi, and Ren Ng.
\newblock Nerf: Representing scenes as neural radiance fields for view
  synthesis.
\newblock {\em Communications of the ACM}, 65(1):99--106, 2021.

\bibitem{muller2003particle}
Matthias M{\"u}ller, David Charypar, and Markus Gross.
\newblock Particle-based fluid simulation for interactive applications.
\newblock In {\em Proceedings of the 2003 ACM SIGGRAPH/Eurographics symposium
  on Computer animation}, pages 154--159. Citeseer, 2003.

\bibitem{muller2022instant}
Thomas M{\"u}ller, Alex Evans, Christoph Schied, and Alexander Keller.
\newblock Instant neural graphics primitives with a multiresolution hash
  encoding.
\newblock {\em ACM Transactions on Graphics (ToG)}, 41(4):1--15, 2022.

\bibitem{IsaacSim}
NVIDIA.
\newblock Isaac sim 4.0 - robotics simulation and synthetic data generation.
\newblock {\em https://developer.nvidia.com/isaac-sim}, 2024.

\bibitem{openai2023gpt4v}
OpenAI.
\newblock {GPT-4V(ision)} system card, 2023.

\bibitem{openai2024gpt4technicalreport}
OpenAI.
\newblock Gpt-4 technical report, 2024.

\bibitem{openai2024gpt4o}
OpenAI.
\newblock Gpt-4o: Openai's optimized language model.
\newblock {\em Technical Report}, 2024.
\newblock Accessed: 2024-10-01.

\bibitem{ouyang2022training}
Long Ouyang, Jeffrey Wu, Xu~Jiang, Diogo Almeida, Carroll Wainwright, Pamela
  Mishkin, Chong Zhang, Sandhini Agarwal, Katarina Slama, Alex Ray, et~al.
\newblock Training language models to follow instructions with human feedback.
\newblock {\em Advances in neural information processing systems},
  35:27730--27744, 2022.

\bibitem{perez2019general}
Diego Perez-Liebana, Jialin Liu, Ahmed Khalifa, Raluca~D Gaina, Julian
  Togelius, and Simon~M Lucas.
\newblock General video game ai: A multitrack framework for evaluating agents,
  games, and content generation algorithms.
\newblock {\em IEEE Transactions on Games}, 11(3):195--214, 2019.

\bibitem{puig2018virtualhome}
Xavier Puig, Kevin Ra, Marko Boben, Jiaman Li, Tingwu Wang, Sanja Fidler, and
  Antonio Torralba.
\newblock Virtualhome: Simulating household activities via programs.
\newblock In {\em Proceedings of the IEEE conference on computer vision and
  pattern recognition}, pages 8494--8502, 2018.

\bibitem{puig2023habitat30cohabitathumans}
Xavier Puig, Eric Undersander, Andrew Szot, Mikael~Dallaire Cote, Tsung-Yen
  Yang, Ruslan Partsey, Ruta Desai, Alexander~William Clegg, Michal Hlavac,
  So~Yeon Min, Vladimír Vondruš, Theophile Gervet, Vincent-Pierre Berges,
  John~M. Turner, Oleksandr Maksymets, Zsolt Kira, Mrinal Kalakrishnan,
  Jitendra Malik, Devendra~Singh Chaplot, Unnat Jain, Dhruv Batra, Akshara Rai,
  and Roozbeh Mottaghi.
\newblock Habitat 3.0: A co-habitat for humans, avatars and robots, 2023.

\bibitem{radford2021learning}
Alec Radford, Jong~Wook Kim, Chris Hallacy, Aditya Ramesh, Gabriel Goh,
  Sandhini Agarwal, Girish Sastry, Amanda Askell, Pamela Mishkin, Jack Clark,
  et~al.
\newblock Learning transferable visual models from natural language
  supervision.
\newblock In {\em International conference on machine learning}, pages
  8748--8763. PMLR, 2021.

\bibitem{raffel2020exploring}
Colin Raffel, Noam Shazeer, Adam Roberts, Katherine Lee, Sharan Narang, Michael
  Matena, Yanqi Zhou, Wei Li, and Peter~J Liu.
\newblock Exploring the limits of transfer learning with a unified text-to-text
  transformer.
\newblock {\em Journal of machine learning research}, 21(140):1--67, 2020.

\bibitem{raistrick2023infinite}
Alexander Raistrick, Lahav Lipson, Zeyu Ma, Lingjie Mei, Mingzhe Wang, Yiming
  Zuo, Karhan Kayan, Hongyu Wen, Beining Han, Yihan Wang, et~al.
\newblock Infinite photorealistic worlds using procedural generation.
\newblock In {\em Proceedings of the IEEE/CVF conference on computer vision and
  pattern recognition}, pages 12630--12641, 2023.

\bibitem{ramesh2022hierarchical}
Aditya Ramesh, Prafulla Dhariwal, Alex Nichol, Casey Chu, and Mark Chen.
\newblock Hierarchical text-conditional image generation with clip latents.
\newblock {\em arXiv preprint arXiv:2204.06125}, 1(2):3, 2022.

\bibitem{ren2024infiniteworldunifiedscalablesimulation}
Pengzhen Ren, Min Li, Zhen Luo, Xinshuai Song, Ziwei Chen, Weijia Liufu, Yixuan
  Yang, Hao Zheng, Rongtao Xu, Zitong Huang, Tongsheng Ding, Luyang Xie,
  Kaidong Zhang, Changfei Fu, Yang Liu, Liang Lin, Feng Zheng, and Xiaodan
  Liang.
\newblock Infiniteworld: A unified scalable simulation framework for general
  visual-language robot interaction, 2024.

\bibitem{shang2024urbanworld}
Yu~Shang, Jiansheng Chen, Hangyu Fan, Jingtao Ding, Jie Feng, and Yong Li.
\newblock Urbanworld: An urban world model for 3d city generation.
\newblock {\em arXiv preprint arXiv:2407.11965}, 2024.

\bibitem{shridhar2020alfred}
Mohit Shridhar, Jesse Thomason, Daniel Gordon, Yonatan Bisk, Winson Han,
  Roozbeh Mottaghi, Luke Zettlemoyer, and Dieter Fox.
\newblock Alfred: A benchmark for interpreting grounded instructions for
  everyday tasks.
\newblock In {\em Proceedings of the IEEE/CVF conference on computer vision and
  pattern recognition}, pages 10740--10749, 2020.

\bibitem{song2023roomdreamer}
Liangchen Song, Liangliang Cao, Hongyu Xu, Kai Kang, Feng Tang, Junsong Yuan,
  and Yang Zhao.
\newblock Roomdreamer: Text-driven 3d indoor scene synthesis with coherent
  geometry and texture.
\newblock {\em arXiv preprint arXiv:2305.11337}, 2023.

\bibitem{sun20233d}
Chunyi Sun, Junlin Han, Weijian Deng, Xinlong Wang, Zishan Qin, and Stephen
  Gould.
\newblock 3d-gpt: Procedural 3d modeling with large language models.
\newblock {\em arXiv preprint arXiv:2310.12945}, 2023.

\bibitem{szot2022habitat20traininghome}
Andrew Szot, Alex Clegg, Eric Undersander, Erik Wijmans, Yili Zhao, John
  Turner, Noah Maestre, Mustafa Mukadam, Devendra Chaplot, Oleksandr Maksymets,
  Aaron Gokaslan, Vladimir Vondrus, Sameer Dharur, Franziska Meier, Wojciech
  Galuba, Angel Chang, Zsolt Kira, Vladlen Koltun, Jitendra Malik, Manolis
  Savva, and Dhruv Batra.
\newblock Habitat 2.0: Training home assistants to rearrange their habitat,
  2022.

\bibitem{geminiteam2024geminifamilyhighlycapable}
Gemini Team.
\newblock Gemini: A family of highly capable multimodal models, 2024.

\bibitem{togelius2011procedural}
Julian Togelius, Emil Kastbjerg, David Schedl, and Georgios~N Yannakakis.
\newblock What is procedural content generation? mario on the borderline.
\newblock In {\em Proceedings of the 2nd international workshop on procedural
  content generation in games}, pages 1--6, 2011.

\bibitem{touvron2023llama2}
Hugo Touvron, Louis Martin, Kevin Stone, Peter Albert, Amjad Almahairi, Yasmine
  Babaei, Nikolay Bashlykov, Soumya Batra, Prajjwal Bhargava, Shruti Bhosale,
  et~al.
\newblock Llama 2: Open foundation and fine-tuned chat models.
\newblock {\em arXiv preprint arXiv:2307.09288}, 2023.

\bibitem{junjue_wang_2021_5706578}
Junjue Wang, Zhuo Zheng, Ailong Ma, Xiaoyan Lu, and Yanfei Zhong.
\newblock Love{DA}: A remote sensing land-cover dataset for domain adaptive
  semantic segmentation, October 2021.

\bibitem{lovada_data}
Junjue Wang, Zhuo Zheng, Ailong Ma, Xiaoyan Lu, and Yanfei Zhong.
\newblock Loveda: A remote sensing land-cover dataset for domain adaptive
  semantic segmentation.
\newblock In J.~Vanschoren and S.~Yeung, editors, {\em Proceedings of the
  Neural Information Processing Systems Track on Datasets and Benchmarks},
  volume~1. Curran Associates, Inc., 2021.

\bibitem{wang2024qwen2}
Peng Wang, Shuai Bai, Sinan Tan, Shijie Wang, Zhihao Fan, Jinze Bai, Keqin
  Chen, Xuejing Liu, Jialin Wang, Wenbin Ge, et~al.
\newblock Qwen2-vl: Enhancing vision-language model's perception of the world
  at any resolution.
\newblock {\em arXiv preprint arXiv:2409.12191}, 2024.

\bibitem{wei2022finetuned}
Jason Wei, Maarten Bosma, Vincent~Y Zhao, Kelvin Guu, Adams~Wei Yu, Brian
  Lester, Nan Du, Andrew~M Dai, and Quoc~V Le.
\newblock Finetuned language models are zero-shot learners.
\newblock {\em arXiv preprint arXiv:2109.01652}, 2021.

\bibitem{worldlabs}
{WorldLabs Team}.
\newblock Worldlabs: Ai-powered virtual world platform.
\newblock \url{https://www.worldlabs.ai/}, 2024.
\newblock Accessed: 2024-12-12.

\bibitem{wu2024metaurban}
Wayne Wu, Honglin He, Jack He, Yiran Wang, Chenda Duan, Zhizheng Liu, Quanyi
  Li, and Bolei Zhou.
\newblock Metaurban: An embodied ai simulation platform for urban
  micromobility.
\newblock {\em arXiv preprint arXiv:2407.08725}, 2024.

\bibitem{wu2021procedural}
Zhixuan Wu, Yuwei Mao, and Qiyu Li.
\newblock Procedural game map generation using multi-leveled cellular automata
  by machine learning.
\newblock In {\em Proceedings of the 2nd International Symposium on Artificial
  Intelligence for Medicine Sciences}, pages 168--172, 2021.

\bibitem{xu2023urban}
Fengli Xu, Jun Zhang, Chen Gao, Jie Feng, and Yong Li.
\newblock Urban generative intelligence (ugi): A foundational platform for
  agents in embodied city environment.
\newblock {\em arXiv preprint arXiv:2312.11813}, 2023.

\bibitem{yi2024gaussiandreamer}
Taoran Yi, Jiemin Fang, Junjie Wang, Guanjun Wu, Lingxi Xie, Xiaopeng Zhang,
  Wenyu Liu, Qi~Tian, and Xinggang Wang.
\newblock Gaussiandreamer: Fast generation from text to 3d gaussians by
  bridging 2d and 3d diffusion models.
\newblock In {\em Proceedings of the IEEE/CVF Conference on Computer Vision and
  Pattern Recognition}, pages 6796--6807, 2024.

\bibitem{text2nerf}
Jingbo Zhang, Xiaoyu Li, Ziyu Wan, Can Wang, and Jing Liao.
\newblock Text2nerf: Text-driven 3d scene generation with neural radiance
  fields.
\newblock {\em arXiv preprint arXiv:2305.11588}, 2023.

\bibitem{zhang2023llama}
Renrui Zhang, Jiaming Han, Chris Liu, Peng Gao, Aojun Zhou, Xiangfei Hu, Shilin
  Yan, Pan Lu, Hongsheng Li, and Yu~Qiao.
\newblock Llama-adapter: Efficient fine-tuning of language models with
  zero-init attention.
\newblock {\em arXiv preprint arXiv:2303.16199}, 2023.

\bibitem{zhang20243d}
Songchun Zhang, Yibo Zhang, Quan Zheng, Rui Ma, Wei Hua, Hujun Bao, Weiwei Xu,
  and Changqing Zou.
\newblock 3d-scenedreamer: Text-driven 3d-consistent scene generation.
\newblock In {\em Proceedings of the IEEE/CVF Conference on Computer Vision and
  Pattern Recognition}, pages 10170--10180, 2024.

\bibitem{zhong2024unrealzooenrichingphotorealisticvirtual}
Fangwei Zhong, Kui Wu, Churan Wang, Hao Chen, Hai Ci, Zhoujun Li, and Yizhou
  Wang.
\newblock Unrealzoo: Enriching photo-realistic virtual worlds for embodied ai,
  2024.

\bibitem{zhou2024scenex}
Mengqi Zhou, Jun Hou, Chuanchen Luo, Yuxi Wang, Zhaoxiang Zhang, and Junran
  Peng.
\newblock Scenex: Procedural controllable large-scale scene generation via
  large-language models.
\newblock {\em arXiv preprint arXiv:2403.15698}, 2024.

\bibitem{zhou2024dreamscene360}
Shijie Zhou, Zhiwen Fan, Dejia Xu, Haoran Chang, Pradyumna Chari, Suya
  Bharadwaj, Tejas~You, Zhangyang Wang, and Achuta Kadambi.
\newblock Dreamscene360: Unconstrained text-to-3d scene generation with
  panoramic gaussian splatting.
\newblock {\em arXiv preprint arXiv:2404.06903}, 2024.

\end{thebibliography}

\end{document}